\pdfoutput=1

\documentclass[11pt]{article}

\usepackage[final]{acl}

\usepackage{times}
\usepackage{latexsym}

\usepackage[T1]{fontenc}

\usepackage[utf8]{inputenc}

\usepackage{microtype}

\usepackage{inconsolata}

\usepackage{graphicx}
\usepackage{amssymb}
\usepackage[table]{xcolor}
\usepackage[normalem]{ulem}

\usepackage{xcolor}
\usepackage[most]{tcolorbox}
\usepackage{booktabs}
\usepackage{makecell}

\title{MPCG: Multi-Round Persona-Conditioned Generation for Modeling the Evolution of Misinformation with LLMs}

\author{Jun Rong Brian Chong \quad 
        Yixuan Tang\thanks{Corresponding author} \quad 
        Anthony K.H. Tung\\[0.5em]
   National University of Singapore  \\[0.25em]
  \texttt{e1327913@u.nus.edu}  \quad  \texttt{\{yixuan, atung\}@comp.nus.edu.sg}
}

\begin{document}
\maketitle
\begin{abstract}
Misinformation evolves as it spreads, shifting in language, framing, and moral emphasis to adapt to new audiences. However, current misinformation detection approaches implicitly assume that misinformation is static. We introduce \textbf{MPCG}, a multi-round, persona-conditioned framework that simulates how claims are iteratively reinterpreted by agents with distinct ideological perspectives. Our approach uses an uncensored large language model (LLM) to generate persona-specific claims across multiple rounds, conditioning each generation on outputs from the previous round, enabling the study of misinformation evolution. We evaluate the generated claims through human and LLM-based annotations, cognitive effort metrics (readability, perplexity), emotion evocation metrics (sentiment analysis, morality), clustering, feasibility, and downstream classification. Results show strong agreement between human and GPT-4o-mini annotations, with higher divergence in fluency judgments. Generated claims require greater cognitive effort than the original claims and consistently reflect persona-aligned emotional and moral framing. Clustering and cosine similarity analyses confirm semantic drift across rounds while preserving topical coherence. Feasibility results show a 77\% feasibility rate, confirming suitability for downstream tasks. Classification results reveal that commonly used misinformation detectors experience macro-F1 performance drops of up to 49.7\%. The code is available at \url{https://github.com/bcjr1997/MPCG}.
\end{abstract}

\section{Introduction}


\begin{figure}[!ht]
    \centering
    \includegraphics[width=0.9\linewidth]{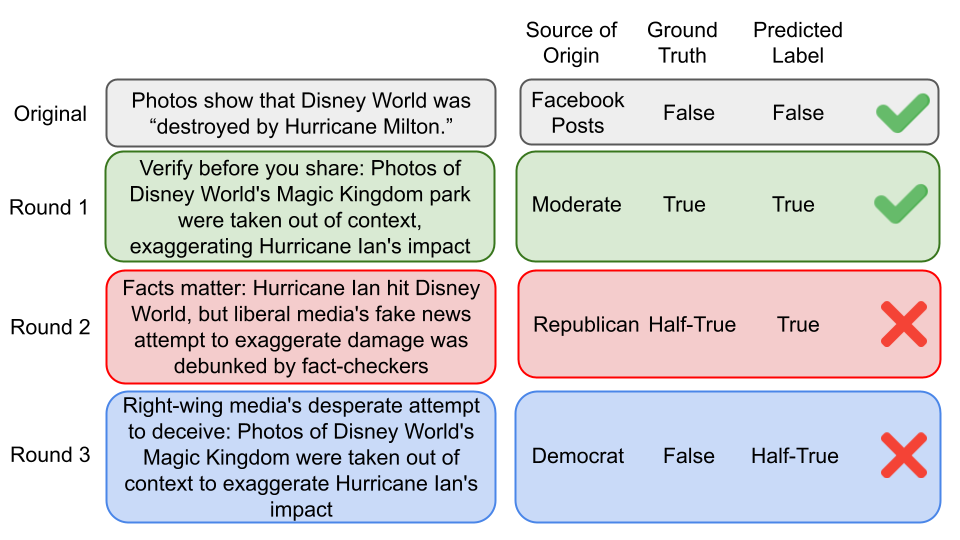}
    \caption{An illustration of how misinformation evolves across perspectives. As each persona reinterprets the original claim, static AFC systems progressively fail to classify the transformed variants, highlighting their limitations against evolving misinformation.}
    \label{fig:motivation}
\end{figure}

Misinformation remains a persistent societal threat, influencing our lives in many ways. Although the Automated Fact Checking (AFC) community has made significant strides through specialized datasets \cite{vlachos-riedel-2014-fact, thorne-etal-2018-fever}, improving explainability \cite{wang-shu-2023-explainable}, and integrating intent features \cite{10.1145/3627673.3679799, DBLP:conf/emnlp/TRACER}, misinformation remains difficult to contain.

While modern misinformation research focuses on verifying the veracity of claims \cite{Simeone_Roschke_Walker_2024}, current AFC approaches face distinct challenges when misinformation evolves. Supervised methods such as fine-tuning rely on fixed claim datasets \cite{wang-2017-liar, thorne-etal-2018-fever, schlichtkrull2023averitec} and cannot handle novel claim variants outside of their training distribution. Advanced LLM-based approaches do mitigate this but remain constrained by knowledge cutoffs, while retrieval-augmented methods can struggle when reliable sources are unavailable.

As shown in Figure~\ref{fig:motivation}, these limitations compound when misinformation evolves. In reality, misinformation is dynamic: it evolves in language, framing, and moral emphasis, evades detection, and continues resonating with target audiences. This evolving nature undermines current AFC methods, making it crucial to model not only static claims but also their potential transformations.

Recent works have explored misinformation generation to pollute data in question answering systems \cite{pan-etal-2023-risk}, annotate claims based on selected evidence \cite{bussotti-etal-2024-unknown}, and disguise fake news by restyling to evade fake news detectors \cite{10.1145/3637528.3671977}. However, these are largely one-shot generation methods and fail to model how misinformation evolves ideologically. Likewise, existing AFC datasets that are derived from fact-checking websites such as PolitiFact\footnote{\url{https://www.politifact.com/}} and Snopes\footnote{\url{https://www.snopes.com/}} focus on verifying individual claims, without annotations capturing their variations tailored to specific audiences.

To our knowledge, limited work has explored how claims evolve through iterative reinterpretations by ideologically distinct personas. Existing LLM-based generation approaches are typically one-shot and lack mechanisms to simulate semantic and stylistic mutation across perspectives. In contrast, persona conditioning provides a structured way to simulate belief-driven reframing, while multi-round generation enables the modeling of misinformation transformation, both critical in understanding how misinformation evolves and persists.

To address this, we propose \textbf{MPCG} (\textbf{M}ulti-round \textbf{P}ersona-\textbf{C}onditioned \textbf{G}eneration), a framework that simulates misinformation evolution by iteratively reframing claims through different ideological personas representing different political stances. In each round, an uncensored LLM generates a new claim based on a target persona, the original claim, and previous generated claims. This cumulative setup enables the modeling of misinformation evolution across different ideological perspectives while maintaining topic coherence.

Our main contributions are:
\begin{itemize}
    \item We formally introduce a new task of multi-round claim generation where LLMs simulate how misinformation dynamically adapts across ideological viewpoints.
	\item We propose an interpretable generation framework that simulates the iterative misinformation transformation through role-playing agents. By conditioning on both prior claims and ideological personas, our method produces realistic, persona-aligned misinformation variants. 
    \item We conduct extensive experiments encompassing human and LLM-based assessments, cognitive and emotional metrics, semantic drift analysis, claim feasibility analysis, and downstream detection robustness. The evaluation results demonstrate the framework’s effectiveness in stress-testing current misinformation detection systems.
\end{itemize}

\section{Related Works}

\subsection{Claim Generation with LLMs}
Claim generation was first introduced as the task of producing claims from information extracted from Wikipedia using human annotators \cite{thorne-etal-2018-fever}. Originally motivated by data scarcity for fact verification, this approach has evolved into automated approaches. Encoder-based transformer models such as BART \cite{lewis-etal-2020-bart} have been used to convert question-answer pairs into claims \cite{pan-etal-2021-zero} and generate scientific claims \cite{wright-etal-2022-generating}. More recently, decoder-based transformer models such as LLMs have been used to generate claims based on selected evidence \cite{bussotti-etal-2024-unknown}. 

However, current claim generation frameworks do not account for how claims can be reinterpreted when expressed by individuals with different backgrounds. As misinformation spreads, each iteration subtly reshapes the structure of the claims while preserving the main topic of the original claim and its sources. Our work addresses this gap by proposing a multi-round, persona-conditioned claim generation framework, where claims are iteratively interpreted and reconstructed by role-playing agents based on their assigned personas and previously generated claims. This design enables us to model how misinformation transforms over time through social reinterpretations.

\subsection{Role-Playing with LLMs}
Role-playing refers to the act of aligning LLMs with specific personas or characters \cite{chen2025oscarsaitheatersurvey} to simulate distinct behaviors or viewpoints. Common implementations include fine-tuning open-source models with role-playing datasets \cite{wang-etal-2024-rolellm} or prompting models with well-crafted profiles, often referred to as personas \cite{wang-etal-2024-unleashing}. In misinformation research, role-playing with LLMs has been used to simulate social media environments. Applications include generating synthetic comments through user-to-user interactions \cite{wan-etal-2024-dell} and simulating the spread of rumors \cite{hu2025simulatingrumorspreadingsocial} and fake news \cite{liu2024skepticismacceptancesimulatingattitude}.

To our knowledge, few studies have applied role-playing specifically for misinformation claim generation. Our work extends this line of research by leveraging multi-round, persona-conditioned generation to analyze how misinformation evolves across different perspectives. This setup provides a structured and interpretable way to study the forms that misinformation takes as it spreads, offering a new direction for misinformation generation research.
\section{Role-Playing Claim Generation Framework}

\begin{figure*}[!ht]
	\centering
    \includegraphics[width=0.78\linewidth]{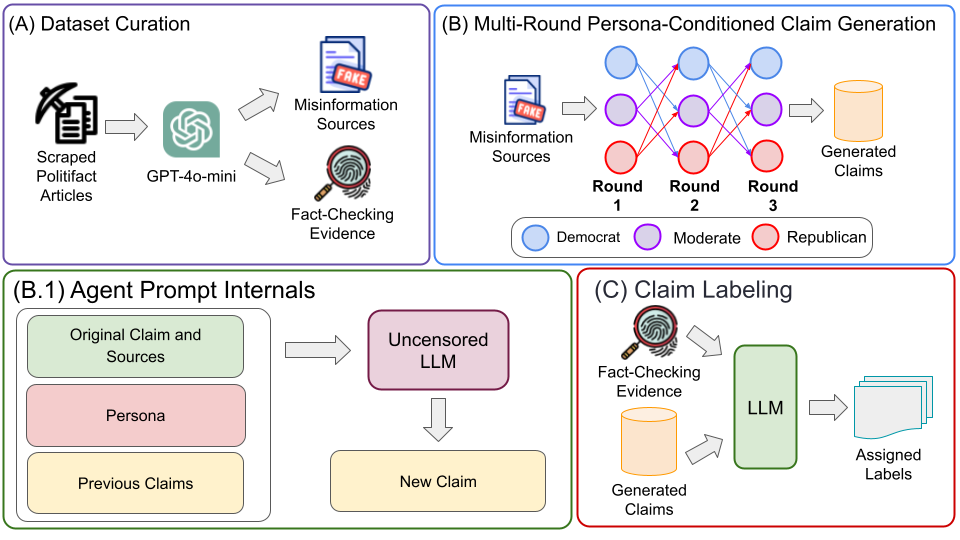}
	\caption{Overview of the \textbf{MPCG} Framework}
	\label{fig:framework}
\end{figure*}

\subsection{Overview}
We introduce \textbf{MPCG}, a framework designed to simulate misinformation evolution for claims. As illustrated in Figure~\ref{fig:framework}, \textbf{MPCG} operates in three stages:
\begin{enumerate}
    \setlength{\itemsep}{0pt}       
    \setlength{\parskip}{0pt}       
    \setlength{\topsep}{0pt}        
    \setlength{\partopsep}{0pt}     
     \setlength{\leftmargin}{1em}    
    \item \textbf{Dataset Curation:} Scrape PolitiFact articles and use GPT-4o-mini to extract both \textbf{Misinformation Sources} and \textbf{Fact-Checking Evidence}.
    \item \textbf{Multi-Round Persona-Conditioned Claim Generation:} Generate persona-aligned claims over three rounds using a structured LLM pipeline, conditioning each generation on the original claim and prior outputs to simulate misinformation evolution.
    \item \textbf{Claim Labeling:} Annotate each generated claim with veracity labels (\texttt{True}, \texttt{Half True}, \texttt{False}) using a structured LLM pipeline for downstream evaluation.
\end{enumerate}

\subsection{Problem Definition}
Given an original claim $C_0$ authored by an individual $CO$, a set of contextual sources $S$, a sequence of personas $P_1, P_2, \ldots, P_k$ where each persona is a tuple $(R_k, D_k)$ representing a role and its description, and optionally a set of previously generated claims $C_{<k}$, the goal is to generate a sequence of claims $C_1, C_2, \ldots, C_k$. Each claim $C_k$ should reflect the viewpoint of persona $P_k$. We define the generation function $G$ as:

\begin{equation}
	C_k = G(C_{<k}, C_0, CO, S, R_k, D_k)
\end{equation}

The function $G$ is implemented using a multi-step prompting pipeline applied to an uncensored LLM, which instructs the model through structured instructions to simulate persona-conditioned interpretation. The generated claims are not necessarily factually accurate, as the personas may introduce bias, exaggeration, or misleading information.

\subsection{Persona Curation and Setup}
\label{sec:role_def_and_cur}
We defined three personas based on the American political spectrum: \texttt{Democrat}, \texttt{Republican} and \texttt{Moderate}. These roles allow us to analyze how claims evolve across different perspectives. For each role, we curated detailed role descriptions to generate claims that align with how these roles are perceived in society. Additional details are described in Appendix \ref{ap:persona_curation}.

\subsection{Dataset Curation}
\label{sec:dataset_curation}
\paragraph{Motivation} To ensure grounded claim generation and evaluation, our framework curates a custom dataset that addresses limitations in existing fact-checking datasets such as  LIAR \cite{wang-2017-liar}, LIAR-PLUS \cite{alhindi-etal-2018-evidence} and AVeriTeC \cite{schlichtkrull2023averitec}, which lack both background context and evidence.

\paragraph{Data Source} We collect 22,408 articles from PolitiFact\footnote{\url{https://www.politifact.com}}, a reputable fact-checking source in English. Each article includes a highlighted claim, background context, evidence, cited sources, and a final veracity label. Table~\ref{tab:raw_dataset_annotations} shows the raw dataset distribution up to 17 March 2025.

\begin{table}[ht]
\small
\centering
\begin{tabular}{lcrlcr}
\toprule
\textbf{Label} & \textbf{Count} & & \textbf{Label} & \textbf{Count} \\
\midrule
True & 2,263 & & Mostly False & 3,407 \\
Half True & 3,431 & & False & 7,010 \\
Mostly True & 3,187 & & Pants on Fire & 3,110 \\
\midrule
\multicolumn{4}{l}{\textbf{Total}} & \textbf{22,408} \\
\bottomrule
\end{tabular}
\caption{Raw data annotation counts from PolitiFact.}
\label{tab:raw_dataset_annotations}
\end{table}

\paragraph{Data Creation} Most articles follow a consistent format: an introduction presenting the claim and its background, a transition sentence marking the beginning of the debunking phase, followed by detailed debunking process and conclude with a final label. Due to the variability in article length and writing style, rule-based extraction proved unreliable.

Inspired by prior work \cite{chatrath2024factfictionllmsreliable}, we used GPT-4o-mini to extract two components from each article using the "Our Sources" section:
\begin{itemize} 
    \setlength{\itemsep}{0pt}       
    \setlength{\parskip}{0pt}       
    \setlength{\topsep}{0pt}        
    \setlength{\partopsep}{0pt}     
     \setlength{\leftmargin}{1em}    
    \item \textbf{Misinformation Sources:} Background context supporting the claim. 
    \item \textbf{Fact-Checking Evidence:} Evidence used to verify or debunk the claim.
\end{itemize}

The complete annotation prompt is provided in Appendix~\ref{ap:dataset_annotation_prompt}

\paragraph{Data Formatting and Verification} The annotated outputs are cleaned and formatted. The PolitiFact veracity labels (\texttt{True, Mostly True, Half True, Mostly False, False, Pants on Fire}) are consolidated into three categories, \texttt{True}, \texttt{Half True} and \texttt{False} as shown in Table~\ref{tab:combined_label_annotations}. This consolidation was necessary as our misinformation detectors could not distinguish subtle label differences during initial testing. 

\begin{table}[ht]
    \small
    \centering
    \begin{tabular}{cccc}
        \toprule
        \textbf{True} & \textbf{Half True} & \textbf{False} & \textbf{Total} \\
        \midrule
        2,263 & 6,618 & 13,527 & 22,408 \\
        \bottomrule
    \end{tabular}
    \caption{Raw data annotations statistics after label combination}
    \label{tab:combined_label_annotations}
\end{table}

\begin{table}[!ht]
\centering
\small
\begin{tabular}{lcc}
\hline
\textbf{Categories} & \textbf{Count} & \textbf{Ratio (\%)} \\
\hline
No Issues & 36 & 72.0 \\
Contaminated Sources & 9 & 18.0 \\
Poor Extraction & 5 & 10.0 \\
\hline
\end{tabular}
\caption{Manual verification results on 50 samples.}
\label{tab:annotation_quality}
\end{table}

To evaluate the annotation quality, we manually reviewed 50 of our annotated samples. Each sample was assessed for extraction accuracy and categorized into one of three groups: \texttt{No Issues}, \texttt{Contaminated Sources}, and \texttt{Poor Extraction}. \texttt{No Issues} indicates satisfactory annotation quality, \texttt{Contaminated Sources} refers to cases where \textbf{Misinformation Sources} contain debunking statements, and \texttt{Poor Extraction} indicates low quality outputs for both \textbf{Misinformation Sources} and \textbf{Fact-Checking Evidence}. In many cases, the extracted content covered only a subset of the listed sources, likely due to the capabilities of GPT-4o-mini. Despite these limitations, the extraction quality for \textbf{Fact-Checking Evidence} was generally accurate. Overall, the annotations were deemed sufficient for our framework. Table \ref{tab:annotation_quality} summarizes the distribution of these findings.

\subsection{Multi-Round Persona-Conditioned Claim Generation}
\label{sec:multi-round_cg_l}
\textbf{MPCG} simulates misinformation evolution by generating persona-specific claims over three rounds using an uncensored LLM as shown in Figure~\ref{fig:framework}. In Round 1, each agent $A$ generates a claim $C_k$ based on the original claim $C_0$, its author $CO$, assigned persona $P$, and contextual sources $S$. In Round 2 and 3, previously generated claims $C_{<k}$ are included to model how misinformation might evolve through reinterpretation by ideologically distinct agents.

Generation is performed using \textbf{Llama-3.1-8B-Lexi-Uncensored-V2}, an uncensored \textbf{LLaMA-3.1-8B-Instruct} model provided by OrengUteng in HuggingFace \footnote{\url{https://huggingface.co/Orenguteng/Llama-3.1-8B-Lexi-Uncensored-V2}}. This model was selected following preliminary experiments with the standard LLaMA-3.1-8B-Instruct model, which frequently rejected
prompts due to its safety alignment mechanisms. Each round is implemented through a structured five-step prompting pipeline executed within a single content window:
\begin{enumerate}
    \setlength{\itemsep}{0pt}       
    \setlength{\parskip}{0pt}       
    \setlength{\topsep}{0pt}        
    \setlength{\partopsep}{0pt}     
     \setlength{\leftmargin}{1em}    
	\item \textbf{Source Reasoning Prompt}: Instructs the model to analyze and reason with the original claim $C_0$, its sources $S$, and available previous claims $C_{<k}$ from the perspective of the assigned persona $P$.
	\item \textbf{Claim Generation Prompt}: Instructs the model to generate a new 20-word claim based on its reasoning.
	\item \textbf{Intent Generation Prompt}: Instructs the model to state its intent when generating the claim.
	\item \textbf{Explanation Prompt}: Instructs the model to generate an explanation based on the claim.
	\item \textbf{Formatting Prompt}: Request the model to provide a response in JSON format.
\end{enumerate}
This prompting design ensures that each generated claim remains topically grounded with the original claim while reflecting the rhetorical and ideological biases of the assigned persona. The prompts can be found in Appendix~\ref{ap:framework_generation_prompts}.

\subsection{Claim Labeling}
To enable downstream classification, each generated claim is assigned a veracity label: \texttt{True, Half True, False} using the provided evidence $E$ as shown in Figure~\ref{fig:framework}. We automate this process using a structured labeling pipeline using \textbf{Llama-3.1-8B-Instruct} \cite{grattafiori2024llama}. The pipeline consists of three prompts executed within a single content window:
\begin{enumerate}
    \setlength{\itemsep}{0pt}       
    \setlength{\parskip}{0pt}       
    \setlength{\topsep}{0pt}        
    \setlength{\partopsep}{0pt}     
     \setlength{\leftmargin}{1em}    
	\item \textbf{Evidence Analysis}: Guides the model in analyzing the generated claim $C_k$ by comparing it with evidence.
	\item \textbf{Label Assignment}: Asks the model to assign the appropriate label and provide a confidence score.
	\item \textbf{Formatting and Label Selection}: Asks the model to select the label with the highest confidence score and return it in JSON format.
\end{enumerate}

All outputs are stored in JSON for our downstream classification task. The prompts can be found in Appendix~\ref{ap:framework_labeling_prompts}.
\section{Experiments}
We evaluated the effectiveness of \textbf{MPCG} by addressing the following key research questions:
\begin{enumerate}
    \setlength{\itemsep}{0pt}       
    \setlength{\parskip}{0pt}       
    \setlength{\topsep}{0pt}        
    \setlength{\partopsep}{0pt}     
    \item \textbf{RQ1: Human-Level Misinformation Quality.} Can \textbf{MPCG} generate persona-conditioned misinformation that aligns with human-level quality in terms of role-playing consistency, content relevance, fluency, factuality, and veracity assignment?
    \item \textbf{RQ2: Linguistic and Moral Characteristics.} What linguistic, emotional, and moral features do the generated claims exhibit, and how do these features evolve across rounds?
    \item \textbf{RQ3: Impact on Classifier Robustness.} How does multi-round claim evolution affect the accuracy and robustness of existing misinformation classifiers?
    \item \textbf{RQ4: Role of Contextual Grounding.} How important are background sources and persona descriptions in shaping the quality and diversity of the generated claims?
\end{enumerate}

\subsection{Dataset}
\label{sec:dataset_final}
To support claim generation and downstream classification tasks, we curate our own dataset as mentioned in Section~\ref{sec:dataset_curation}. Table~\ref{tab:final_data_distribution} presents the dataset statistics after label consolidation and class balancing. The final dataset contains 6,789 samples, evenly distributed across three classes: \texttt{True, Half True, False}. The dataset is split into \textbf{Train, Dev, Test} with 80\%/10\%/10\% ratio. The \texttt{Test} set is used for claim generation in our framework, while the \texttt{Train} and \texttt{Dev} sets are used to finetune the encoder-based misinformation classifiers. 

\begin{table}[ht]
    \centering
    \small
    \begin{tabular}{ccccc}
        \toprule
        \textbf{Dataset Type} & \textbf{True} & \textbf{Half True} & \textbf{False} & \textbf{Total} \\
        \midrule
        Train & 1,811 & 1,811 & 1,811 & 5,433 \\
        Dev & 226 & 226 & 226 & 678 \\
        Test & 226 & 226 & 226 & 678 \\
        \bottomrule
    \end{tabular}
    \caption{Final dataset distribution after label consolidation and balancing}
    \label{tab:final_data_distribution}
\end{table}

\vspace{-1.0em}
\subsection{Evaluation Setup and Metrics}
\label{sec:experiments}
All experiments are performed using an NVIDIA A100 GPU on Google Colab and GPT-4o-mini. Generating 10,170 claims and labeling them took about 2 days. We evaluate the generated claims using three complementary evaluations.

\paragraph{Human and GPT-4o-mini Evaluation} 
We conduct a questionnaire-based evaluation with 30 university graduates familiar with American politics. Annotators rate the generated claims based on role-playing consistency, content relevance, fluency, and factuality using a 5-point Likert scale. They are tasked with assigning veracity labels \texttt{(True, Half True, False)} based on the provided evidence. The same task is performed with GPT-4o-mini. To quantify agreement between human and GPT-4o-mini responses across all rating dimensions, we compute the binned Jensen-Shannon Divergence (JSD) \cite{menendez1997jensen,  elangovan2025correlationimpacthumanuncertainty} using \texttt{jensenshannon} provided by SciPy \cite{2020SciPy-NMeth}. Additional details are provided in Appendix~\ref{ap:human_eval}.

\paragraph{Claim Analysis} We analyze the linguistic and emotional features of the generated claims using metrics from previous work \cite{natureFingerprintsMisinformation}. These metrics are grouped into three categories: \textbf{Cognitive Effort}, \textbf{Emotion Evocation}, and \textbf{Clustering}. \textbf{Cognitive Effort} measures the processing difficulty of the claim using readability and perplexity \cite{natureFingerprintsMisinformation}. We measure readability using Flesch-Kincaid Grade Level (FKGL) score \cite{kincaid1975derivation} via TextStat\footnote{\url{https://pypi.org/project/textstat/}} and perplexity using GPT-2 via HuggingFace\footnote{\url{https://huggingface.co/docs/transformers/en/perplexity}}. Perplexity indirectly measures lexical diversity through text quality \cite{tevet-berant-2021-evaluating}. \textbf{Emotion Evocation} measures the emotional appeal of the claim using sentiment analysis and morality \cite{natureFingerprintsMisinformation}. Sentiment is measured via the \texttt{sentiment-analysis} pipeline provided by HuggingFace\footnote{\url{https://huggingface.co/blog/sentiment-analysis-python}} using \texttt{cardiffnlp/twitter-roberta-base-sentiment} \cite{barbieri-etal-2020-tweeteval}. We measure Morality using MoralBERT \cite{10.1145/3677525.3678694} which measures the morality of a given text based on ten moral foundations. Additional morality details are provided in Appendix~\ref{ap:claim_analysis}. \textbf{Clustering} measures the semantic deviations between the generated claims and their original claims using \texttt{all-MiniLM-L6-v2} model provided by SBERT \cite{reimers-2019-sentence-bert}, HDBSCAN \cite{Malzer_2020} with \texttt{min\_cluster\_size} of 5 and UMAP \cite{McInnes2018} with a random state of 42 in its default settings. \textbf{Feasibility} measures how feasible a claim is to assess for veracity \cite{10.1145/3711896.3737437}. We use the Evaluation Quality Assurance (EQA) tool provided by the original authors to measure the feasibility of our generated claims in its default settings with \texttt{gpt-4o-mini-2024-07-18} as the evaluator model.

\paragraph{Classification} 
We evaluate the impact of our generated claims on downstream tasks using classification with commonly used encoder-based and decoder-based models in misinformation detection.
To ensure fair comparison across all generated claims derived from their original sources, we reuse the gold evidence originally used by PolitiFact to debunk the original claim as the basis for evaluating its generated variants. This controlled setup isolates the impact of claim evolution on classifier performance and enables robustness evaluation. We measure the macro precision, recall, and F1 scores using \texttt{precision\_score, recall\_score, f1\_score} provided by Scikit-Learn \cite{scikit-learn}. For encoder-based models, we fine-tune BERT \cite{devlin-etal-2019-bert}, RoBERTa \cite{liu2019robertarobustlyoptimizedbert} and DeBERTa-v3 \cite{he2023debertav3improvingdebertausing} and their large variants using the HuggingFace Trainer\footnote{\url{https://huggingface.co/docs/transformers/en/main_classes/trainer}} framework and our training dataset stated in Section~\ref{sec:dataset_final}. For decoder-based models, we use LLaMA 3.1-8B Instruct and GPT-4o-mini to label these claims via zero-shot, few-shot, zero-shot chain-of-thought (CoT) \cite{10.5555/3600270.3602070} and few-shot CoT prompting strategies. The prompts, fine-tuning approaches, and additional details are provided in Appendix~\ref{ap:classification}.

\paragraph{Ablation Studies} We conduct ablation studies to evaluate the robustness of our framework by removing key input components. Specifically, we assess the impact of removing \texttt{Role Descriptions} and omitting \texttt{Background Sources} from the generation prompts. We evaluate these ablations using \texttt{GPT-4o-mini} tested with \texttt{Role-Playing Consistency} and \texttt{Content Relevance} prompts shown in Appendix~\ref{ap:human_eval_rp} and ~\ref{ap:human_eval_relevance} respectively. These experiments help isolate the contribution of contextual and persona grounding to the quality and variations of generated claims.
\section{Results and Discussion}
We evaluate \textbf{MPCG} across the following stages: \textbf{Original} refers to the original PolitiFact claims; \textbf{Round 1, 2, 3} correspond to the generated claims using our framework in Figure~\ref{fig:framework}. 

\subsection{Human and GPT-4o-mini Evaluation}
\begin{table}
    \small
    \centering
    \begin{tabular}{lcc}
        \toprule
        \textbf{Question} &  \textbf{JSD}\\
        \midrule
        Role-Playing Consistency (Q1) & 0.178\\
        Content Relevance (Q2) & 0.174\\
        Fluency (Q3) & 0.296\\
        Factuality (Q4) & 0.195\\
        Label Assignment (Q5) & 0.113\\
        \bottomrule
    \end{tabular}
    \caption{Jensen Shannon Divergence (JSD) scores for 363 human and GPT-4o-mini evaluations. Lower is better.}
    \label{tab:jsd_human_eval}
\end{table}

To assess whether the generated claims resemble human-level misinformation, we examine the alignment between human ratings and GPT-4o-mini evaluations. Table~\ref{tab:jsd_human_eval} shows strong alignment between human and GPT-4o-mini ratings for most dimensions, with higher divergence in fluency. GPT-4o-mini favors "Excellent" while humans tend to select "Good", reflecting a potential judgment bias \cite{chen-etal-2024-humans}. Despite this, both rate fluency highly overall, indicating general fluency of generated claims.

\subsection{Claim Analysis}
\begin{table}[!h]
\small
\centering
\begin{tabular}{cccccc}
\toprule
\textbf{Persona} & \textbf{Round} & \textbf{Med.} & \textbf{Q1} & \textbf{Q3} & \textbf{IQR} \\
\midrule
Original & -- & 9.10  & 6.90 & 11.9 & 5.00 \\
Democrat    & 1 & 14.0 & 12.0 & 16.1 & 4.10 \\
Moderate    & 1 & 14.0 & 12.3 & 16.0 & 3.70 \\
Republican  & 1 & 14.4 & 12.4 & 16.2 & 3.80 \\
Democrat    & 2 & 14.5 & 12.5 & 16.7 & 4.20 \\
Moderate    & 2 & 14.5 & 12.4 & 16.4 & 4.00 \\
Republican  & 2 & 14.8 & 12.9 & 16.7 & 3.80 \\
Democrat    & 3 & 14.9 & 12.9 & 16.8 & 3.90 \\
Moderate    & 3 & 14.9 & 12.8 & 17.0 & 4.20 \\
Republican  & 3 & 14.6 & 12.8 & 16.8 & 4.00 \\
\bottomrule
\end{tabular}
\caption{Flesch-Kincaid Grade Level (FKGL) scores for original and generated claims across all rounds. Higher scores indicate more syntactically complex text.}
\label{tab:readability}
\end{table}

\begin{table}
\small
\centering
\begin{tabular}{cccccc}
\toprule
\textbf{Persona} & \textbf{Round} & \textbf{Med.} & \textbf{Q1} & \textbf{Q3} & \textbf{IQR} \\
\midrule
Original & -- & 57.0 & 32.8 & 107.2 & 74.4 \\
Democrat & 1 & 45.3 & 29.9 & 72.6 & 42.7 \\
Moderate & 1  & 51.5 & 34.8 & 76.2 & 41.4 \\
Republican & 1  & 47.5 & 31.0 & 71.1 & 40.1 \\
Democrat & 2  & 42.4 & 29.7 & 66.3 & 36.6 \\
Moderate & 2  & 50.2 & 35.5 & 74.9 & 39.4 \\
Republican & 2  & 48.1 & 33.4 & 75.3 & 41.9 \\
Democrat & 3  & 44.7 & 30.9 & 71.0 & 40.1 \\
Moderate & 3  & 49.8 & 34.5 & 79.3 & 44.8 \\
Republican & 3  & 48.8 & 33.8 & 74.7 & 40.9 \\
\bottomrule
\end{tabular}
\caption{Perplexity scores for original and generated claims across all rounds. Lower scores indicate less lexical diversity and higher word-level predictability.}
\label{tab:perplexity}
\end{table}

\paragraph{Cognitive Effort} To characterize the linguistic properties of generated claims, we analyze their readability and lexical diversity. Table~\ref{tab:readability} shows that generated claims are syntactically more complex than the original. Table~\ref{tab:perplexity} shows less lexically diverse when compared to the original. Taken together, these results suggest that generated claims require a higher education level to comprehend while relying on a more consistent vocabulary set.

\begin{table}
\small
\centering
\begin{tabular}{ccccc}
\toprule
\textbf{Persona} & \textbf{Round} & \textbf{Negative} & \textbf{Neutral} & \textbf{Positive} \\
\midrule
Original      & -              & 281               & \textbf{350}              & 47               \\
Democrat      & 1              & \textbf{365}               & 179              & 134               \\
Moderate      & 1              & 152               & \textbf{399 }             & 127               \\
Republican    & 1              & \textbf{319}               & 195              & 164               \\
Democrat      & 2              & \textbf{598}               & 435              & 323               \\
Moderate      & 2              & 300               & \textbf{759}              & 297               \\
Republican    & 2              & \textbf{540}               & 403              & 413               \\
Democrat      & 3              & \textbf{547 }              & 436              & 373               \\
Moderate      & 3              & 222               & \textbf{760 }             & 374               \\
Republican    & 3              & \textbf{491 }              & 457              & 408               \\
\bottomrule
\end{tabular}
\caption{Distribution of sentiments for original and generated claims across rounds and personas. Bolded values indicate the majority sentiment class within each group.}
\label{tab:sentiment_distribution}
\end{table}

\begin{table}
\small
\centering
\begin{tabular}{cccc}
\toprule
\textbf{MFT} & \makecell{\textbf{Persona}\\\textbf{(Round)}} & \makecell{\textbf{Avg}\\\textbf{Score}} & \textbf{SE} \\
\midrule
Authority & Republican (Round 3) &  0.106 & 0.007 \\
Betrayal & Democrat (Round 2)   &  0.048 & 0.004 \\
Care & Democrat (Round 3)   &  0.183 & 0.006 \\
Cheating & Democrat (Round 1)   &  0.183 & 0.012 \\
Degradation & Original &  0.072 & 0.004 \\
Fairness & Democrat (Round 3) &  0.264 & 0.011 \\
Harm & Original &  0.109 & 0.009 \\
Loyalty & Moderate (Round 3)   &  0.018 & 0.003 \\
Purity & Original &  0.007 & 0.003 \\
Subversion & Republican (Round 3) &  0.013 & 0.002 \\
\bottomrule
\end{tabular}
\caption{Morality scores across original and generated claims based on Moral Foundations Theory (MFT). Higher scores indicate stronger moral framing expressed in the generated claims.}
\label{tab:morality}
\end{table}

\paragraph{Emotion Evocation} We next analyze whether the generated claims reflect the emotional and moral orientations of their assigned personas. Table~\ref{tab:sentiment_distribution} shows Democrat and Republicans produce more negative claims, while Moderates remain mostly neutral. Table~\ref{tab:morality} shows Democrats emphasize care, fairness, cheating, and betrayal, while Republicans emphasize authority and subversion, aligning with previous work where liberals rely heavily on harm and fairness while conservatives rely more on authority \cite{day2014shifting}. In contrast, Moderates generate more neutral claims while emphasizing loyalty, likely reflecting a downplaying or a neutralizing strategy during the generation process. These results indicate that our framework can generate claims that are morally and sentimentally aligned with the personas, poised to resonate with its intended audience.

\begin{figure}
	\centering
	\includegraphics[width=0.8\linewidth]{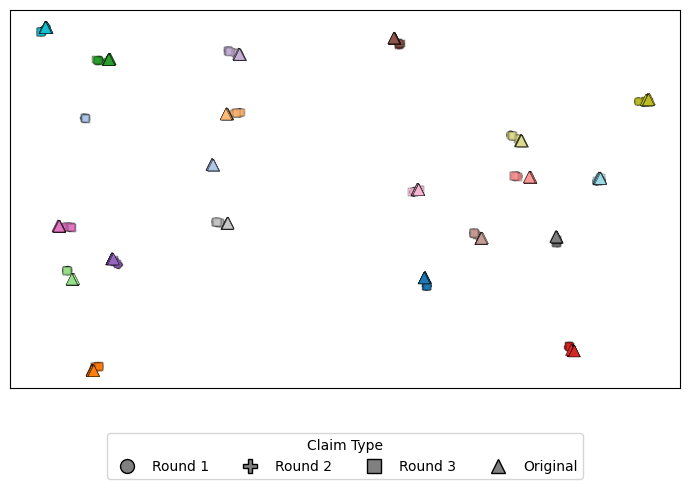}
	\caption{Semantic clusters of Round 1 generated claims (Circle), Round 2 generated claims (Plus), Round 3 generated claims (Square), and the original claims (Triangle). Each color represents a group of claims associated with the same original PolitiFact URL.}
	\label{fig:clustering}
\end{figure}

\paragraph{Clustering} To assess whether generated claims preserve their original topics while undergoing stylistic changes, we perform clustering analysis. Figure~\ref{fig:clustering} shows the clusters formed by a sample of 300 claims using SBERT, UMAP and HDBSCAN. Each color corresponds to a unique PolitiFact URL, and each shape represents a different generation round. Our clustering results show that most generated claims remain semantically close to their respective original claims, suggesting that the main topic is preserved as framing shifts. However, a subset of generated claims deviates significantly from their original claims due to the shifts in framing. These results indicate that misinformation can evolve stylistically, changing its tone, emphasis and perspective while preserving topic alignment.

\paragraph{Feasibility} To assess the generated claims are suitable for downstream tasks, we evaluate the feasibility. We evaluated 300 claims based on 20 unique PolitiFact articles using the EQA tool. 231 claims (77\%) were classified as feasible: 203 as "Feasible with web search" and 28 as "Feasible". Our results are close to the feasibility threshold 75\% proposed in the original work \cite{10.1145/3711896.3737437}. However, the high proportion of claims requiring web search indicates that key information are missing, preventing detectors from determining the claim's truthfulness without retrieving additional evidence online. In general, these results indicate that MPCG is capable of generating feasible claims that can be classified for veracity.

\begin{table*}[!htbp]
\small
\centering
\resizebox{\textwidth}{!}{%
\begin{tabular}{lcccccccccccc}
\toprule
\textbf{Model} & \multicolumn{3}{c}{\textbf{Original}} & \multicolumn{3}{c}{\textbf{Round 1}} & \multicolumn{3}{c}{\textbf{Round 2}} & \multicolumn{3}{c}{\textbf{Round 3}} \\
\cline{2-13}
 & P (\%) & R (\%) & F1 (\%) & P (\%) & R (\%) & F1 (\%) & P (\%) & R (\%) & F1 (\%) & P (\%) & R (\%) & F1 (\%) \\
\midrule
BERT$_{Base}$ & 64.3 & 64.6 & 64.4 & 33.2 & 33.3 & 33.3 & 37.4 & 37.6 & 37.4 & 36.3 & 36.4 & 36.3 \\
BERT$_{Large}$ & 67.3 & 65.5 & 65.8 & 36.7 & 36.3 & 35.9 & 38.5 & 38.3 & 37.9 & 36.5 & 35.7 & 35.4 \\
\midrule
RoBERTa$_{Base}$ & 67.8 & 66.7 & 67.0 & 35.6 & 35.1 & 35.0 & 38.7 & 37.9 & 37.8 & 37.5 & 36.7 & 36.7 \\
RoBERTa$_{Large}$ & 71.3 & 70.5 & 70.8 & 36.7 & 36.3 & 36.1 & 39.5 & 38.6 & 38.6 & 39.3 & 38.1 & 38.3 \\
\midrule
DeBERTa V3$_{Base}$ & 70.2 & 69.5 & 69.7 & 39.4 & 39.1 & 38.4 & 37.1 & 36.7 & 36.4 & 37.9 & 36.9 & 36.8 \\
DeBERTa V3$_{Large}$ & \cellcolor{green!20}\textbf{72.9} & \cellcolor{green!20}\textbf{71.2} & \cellcolor{green!20}\textbf{71.7} & 38.7 & 37.5 & 36.1 & 39.6 & 38.6 & 37.5 & 41.0 & 38.6 & 37.8 \\
\midrule
LLaMA 3.1 8B Instruct (Zero Shot) & 62.4 & 61.7 & 61.2 & \cellcolor{green!20}\textbf{44.5} & \cellcolor{green!20}\textbf{44.8} & \cellcolor{green!20}\textbf{44.4} & \cellcolor{green!20}\textbf{46.1} & \cellcolor{green!20}\textbf{46.2} & \cellcolor{green!20}\textbf{45.8} & \cellcolor{green!20}\textbf{43.5} & \cellcolor{green!20}\textbf{43.5} & \cellcolor{green!20}\textbf{43.0} \\
LLaMA 3.1 8B Instruct (Zero Shot CoT) & 61.4 & 54.7 & 51.7 & 49.7 & 45.0 & 42.9 & 45.7 & 43.5 & 41.4 & 42.6 & 39.8 & 37.0 \\
LLaMA 3.1 8B Instruct (Few Shot) & 61.2 & 56.9 & 56.0 & 42.9 & 42.3 & 39.7 & 45.4 & 44.5 & 41.7 & 42.9 & 40.7 & 38.3 \\
LLaMA 3.1 8B Instruct (Few Shot CoT) & 61.5 & 53.1 & 50.6 & 44.1 & 41.2 & 37.0 & 43.3 & 42.3 & 38.1 & 43.8 & 40.4 & 35.4 \\
\midrule
GPT-4o-mini (Zero Shot) & 69.1 & 68.0 & 68.4 & 46.8 & 42.3 & 41.8 & 50.9 & 44.1 & 43.3 & 48.9 & 42.9 & 41.3 \\
GPT-4o-mini (Zero Shot CoT)  & 71.1 & 65.2 & 65.5 & 46.9 & 39.8 & 36.8 & 52.7 & 43.7 & 41.6 & 50.8 & 40.7 & 37.6 \\
GPT-4o-mini (Few Shot)  & 71.8 & 68.4 & 69.1 & 45.2 & 39.4 & 37.1 & 50.0 & 39.8 & 37.4 & 50.7 & 40.6 & 38.1 \\
GPT-4o-mini (Few Shot CoT)  & 72.8 & 67.0 & 67.6 & 48.0 & 40.0 & 36.6 & 49.1 & 40.0 & 37.0 & 50.1 & 39.7 & 36.3 \\
\bottomrule
\end{tabular}%
}
\caption{Macro Average Precision (P), Recall (R) and F1 scores for each model across grouped claims: Original, Round 1, Round 2, and Round 3. Green cells indicate highest macro F1 within each group of claims.}
\label{tab:classification_metrics}
\end{table*}

\begin{table*}
    \small
    \centering
    \begin{tabular}{ccccc}
        \toprule
        \textbf{Configuration} & \textbf{Evaluation} & \textbf{Background Sources} & \textbf{Role Definitions} & \textbf{Weighted Average Score} \\
        \midrule
        1 & Content Relevance        & Y & Y & \cellcolor{green!20}\textbf{3.54} \\
        2 & Content Relevance        & N & Y & 3.40 \\
        \midrule
        1 & Role-Playing Consistency & Y & Y & \cellcolor{green!20}\textbf{4.63} \\
        2 & Role-Playing Consistency & Y & N & 4.56 \\
        \bottomrule
    \end{tabular}
    \caption{Ablation results for contextual grounding for 270 generated claims. Green cells indicate highest weighted average score within each evaluation configuration.}
    \label{tab:ablation_studies}
\end{table*}

\subsection{Classification Robustness}

To evaluate how these generated claims impact common misinformation detectors, we proceed with classification. Table~\ref{tab:classification_metrics} shows that DeBERTa V3$_{Large}$ achieves the highest macro F1 score on original claims, but all models suffer significant drops on generated claims: up to 35.6 percentage points (49.7\%) for encoder-based models and up to 32.0 percentage points (46.3\%) for decoder-based models. This aligns with prior findings that stylistic perturbations can reduce fake news detectors F1 score performance up to 38\% \cite{10.1145/3637528.3671977}.

\begin{table}
    \small
    \centering
    \begin{tabular}{lcc}
        \toprule
        \textbf{Datasets} & \textbf{Count} & \textbf{Avg Cosine Similarity} \\
        \midrule
        Original $\rightarrow$ Round 1 & 2,034 & 0.644 \\
        Original $\rightarrow$ Round 2 & 4,068 & 0.625 \\
        Original $\rightarrow$ Round 3 & 4,068 & 0.618 \\
        Round 1 $\rightarrow$ Round 2 & 4,068 & 0.699 \\
        Round 2 $\rightarrow$ Round 3 & 4,068 & 0.698 \\
        \bottomrule
    \end{tabular}
    \caption{Average Cosine Similarities for each dataset combinations}
    \label{tab:avg_cosine_similarity}
\end{table}

To better understand this trend, we analyze the average cosine similarities in Table~\ref{tab:avg_cosine_similarity}. Average cosine similarities with the original claims steadily decline from Round 1 to Round 3, reflecting incremental semantic drift. However, adjacent rounds (Round 1 $\rightarrow$ and Round 2, and Round 2 $\rightarrow$ and Round 3) maintain high similarity, suggesting that each generation introduces only small shifts, consistent with our clustering results shown in Figure~\ref{fig:clustering}. This explains why classifiers collapse most severely when we move from original claims to \texttt{Round 1}, but then stabilize in later rounds. The plateau in F1 scores reflect a limited form of robustness, as detectors still perform consistently when the variants still revolve around the same underlying topic.

\subsection{Role of Contextual Grounding}

To quantify the importance of \texttt{Background\- Sources} and \texttt{Role\- Definitions} for MPCG, Table~\ref{tab:ablation_studies} reports the ablation results on 270 generated claims. Performance drops from 3.54 to 3.40 and 4.63 to 4.56 when \texttt{Background Sources} and \texttt{Role Definitions} are removed. Although the difference is modest, the results indicate that providing additional context contributes to higher quality claim generation.


\section{Conclusion and Future Work}
In this work, we introduce a new task: multi-round claim generation, where LLMs simulate how misinformation dynamically adapts across ideological viewpoints. We propose \textbf{MPCG}, a novel framework that models misinformation evolution through iterative generation with different ideological personas, capturing how claims evolve.

Human and GPT-4o-mini evaluations show strong alignment in role-playing consistency, relevance, fluency, and factuality, indicating that the generated claims can mimic human level misinformation. Claim analysis reveals that the generated claims require higher cognitive effort and exhibit persona-aligned sentiment and moral framing, suggesting their potential to influence targeted audiences. Clustering and cosine similarity analyses further confirm that claims evolve stylistically and semantically over rounds, while remaining anchored to the same underlying topic. Feasibility analysis shows that most generated claims are verifiable, although many require retrieval of external evidence, underscoring the effectiveness of MPCG in producing feasible verifiable claims.

Classification results indicate that standard misinformation detectors suffer performance degradation on Round 1 claims, with performance plateau in later rounds. This highlights the models' robustness to semantically similar claims and vulnerability to stylistic shifts.

These findings emphasize the evolving nature of misinformation and the importance of modeling its progression. Our framework provides a foundation for stress-testing AFC systems, similar to load testing in software engineering, to help develop more resilient detection methods.

Future work includes developing standardized automatic metrics to evaluate generation quality, reducing reliance on subjective human and LLM assessments prone to judgment bias \cite{chen-etal-2024-humans} and human uncertainty \cite{elangovan2025correlationimpacthumanuncertainty}. Additionally, extending this framework to multilingual and multimodal settings can broaden its applicability and provide insights into how misinformation evolves in different settings.
\section*{Limitations}
While MPCG provides insights into misinformation evolution, it faces several limitations. Dataset quality depends on PolitiFact articles and GPT-4o-mini annotations, which may introduce inaccuracies. Annotation quality remains sensitive to prompt design and model choice, affecting reproducibility across different setups. Additionally, curated personas are based on 2021 typologies and may not reflect current ideological trends, and designing non-generic personas without such typologies remains challenging. Results also depend on a single uncensored LLM and may vary with different uncensored models using the same configurations. 

Furthermore, human evaluation is limited by few annotators and questionnaire design, while classification depends on LLM-generated labels rather than gold-standard expert labels. Moreover, generated claims rely on original PolitiFact evidence for classification, which may not reflect real-world fact-checking scenarios. Although MPCG achieved a 77\% feasibility score, above the proposed 75\% threshold, a moderate degree of noise remains \cite{10.1145/3711896.3737437}. 

The claim analysis metrics present additional limitations. Flesch-Kincaid Grade Level is simplistic and can be exploited \cite{tanprasert-kauchak-2021-flesch}, perplexity loosely reflects lexical diversity, and sentiment and morality analysis rely on subjective classifiers which can introduce bias. 

Finally, although we focus on evaluating politically evolved claims, future work should extend MPCG to test for novel claims or misinformation from different domains, such as healthcare, to examine generalization beyond politics. Some generated outputs may also resemble normative or opinionated claims rather than verifiable claims. This ambiguity is a known challenge in misinformation datasets and could be mitigated in future work with better protocols to distinguish between normative and factual claims.

\section*{Ethical Considerations}
This work involves generating synthetic misinformation content using uncensored LLMs, which presents some ethical risks. While the objective is to study how misinformation evolves, we acknowledge that generating such content may be misused and cause intended harm. To mitigate this, we do not release any generated claims. Instead, we provide the generation code and framework configuration, allowing researchers to replicate the methodology under controlled settings. 

Our persona definitions are based on publicly available sources and are intended to reflect real-world discourse, not to reinforce stereotypes. However, these personas are U.S centric and derived from typologies established in 2021 which may not reflect the current state of communities that share the same typologies. As such, the findings of our study should not be generalized to non-U.S perspectives.

The dataset used contains only publicly available information disclosed in PolitiFact articles. No personally identifiable information is included, and we do not infer any protected attributes such as race, gender, and ethnicity. All data is used for non-commercial academic research.

The outputs of our framework are synthetic and not intended for public deployment. Nonetheless, there is a potential for unintended misuse, including the interpretation of generated claims as real texts. Researchers may find such tools useful for modeling the evolution of misinformation but should exercise cautious in avoiding reinforcing false narratives.

\section*{Acknowledgments}
This research is supported by the Ministry of Education, Singapore, under its MOE AcRF TIER 3 Grant (MOE-MOET32022-0001).

\section*{License and Terms of Use} 
The PolitiFact data used in this work was collected from publicly accessible fact-check articles through our custom web scraper. We acknowledge the terms of use provided by PolitiFact \footnote{\url{https://www.politifact.com/copyright/}} and confirm that the data was collected solely for non-commercial academic research purposes.

\bibliography{anthology,custom}

\onecolumn
\appendix
\section{Persona Curation}
\label{ap:persona_curation}
To ensure that each persona reflects a socially grounded and ideologically representative viewpoint, we establish role descriptions based on multiple sources. 

For \texttt{Democrat} and \texttt{Republican} role descriptions, their definitions were derived from the Wikipedia version of Beyond Red vs. Blue: The Political Typology 2021 from Pew Research Center \footnote{\url{https://en.wikipedia.org/wiki/Pew_Research_Center_political_typology}}, which describes the American political spectrum in 2021 modeled by Pew Research Center. We specifically referenced the Democratic Coalition and the Republican Coalition typologies for their definitions. Due to the variability of the length of our claim's sources, these references were summarized to respect token limitations. Additionally, we incorporate the definitions from Merriam-Webster \footnote{\url{https://www.merriam-webster.com/}} which provides a high-level definition of the role.

The definition for \texttt{Moderate} is derived from Wikipedia's Political Moderate page \footnote{\url{https://en.wikipedia.org/wiki/Political_moderate}} which provides a high-level definition for this role. Furthermore, we also incorporate the characteristics of a \texttt{Moderate} from American Political Science Review \cite{fowler2023moderates} which offers a scholarly perspective for this role. 

The final role descriptions used for our framework are listed below:

\subsection{Democrat}
\begin{itemize}
    \item Younger liberal voters that are skeptical of the political system and both major political parties. They believe that the American political system unfairly favors powerful interests, and about half say that the government is wasteful and inefficient. They are more likely to say that no political candidate represents their political views and least likely to say that there is a "great deal of difference" between the parties.
    \item Older voters that are economically liberal and socially moderate who support higher taxes and expansion of the social safety net as well as stronger military policy. They also see violent crime as a "very big" national problem, to oppose increased immigration, and to say that people being too easily offended is a major problem.
    \item Highly liberal voters who are loyal to the Democratic Party and are more likely than other groups to seek compromise and to hold an optimistic view of society.
    \item Younger highly liberal voters who believe that the scope of government should "greatly expand" and that the institutions of the United States need to be "completely rebuilt" to combat racism. They are the most likely group to say that there are countries better than the United States, that the American military should be reduced, that fossil fuels should be phased out, and that the existence of billionaires is bad for society.
    \item A member of one of the two major political parties in the U.S. that is usually associated with government regulation of business, finance, and industry, with federally funded educational and social services, with separation of church and state, with support for abortion rights, affirmative action, gun control, and policies and laws that protect and support the rights of workers and minorities, and with internationalism and multilateralism in foreign policy.
\end{itemize}

\subsection{Republican}
\begin{itemize}
    \item Highly conservative and highly religious voters who generally support school prayer and military over diplomacy while generally oppose legalized abortion and same-sex marriage. They are more likely to claim that the United States "stands above all other countries in the world" and that illegal immigration is a "very big national problem", known to be staunch pro-Israel supporters, are more likely to reject the concept of white privilege and to agree that white Americans face more discrimination than African Americans and people of color.
    \item Conservative voters that emphasize pro-business views, international trade and small government who hold moderate views on immigration and race than other groups within the Republican coalition.
    \item Highly conservative anti-immigrant voters that oppose the role of government and big businesses in American society. They are more likely to believe that the number of legal immigrants should decrease and that the decreasing proportion of white Americans is bad for society. They are also more likely to support raising taxes on the rich.
    \item Younger voters that lean conservative on economic and race issues but lean moderate on social issues. They are more likely to support diplomacy over military strength, legalized marijuana, legalized abortion and "openness to people from all over the world".
    \item A member of one of the two major political parties in the United States that is usually associated with reduced taxation, with limited government regulation of business, finance, industry, education, and policing, with strong national defense, and with opposition to abortion, affirmative action, gun control, and policies and laws that are viewed as challenging traditional social and family hierarchies and structure.
\end{itemize}

\subsection{Moderate}
\begin{itemize}
    \item An ideological category which designates a rejection of radical or extreme views, especially in regard to politics and religion. 
    \item Someone occupying any mainstream position to avoid extreme views. 
    \item Often described as politically unsophisticated, uninformed, or ideologically innocent, secretly partisan, ideologically cross-pressured, or extreme, with patterns of attitudes poorly described by a single ideological dimension
\end{itemize}

\section{Claim Generation Prompt Templates}
\label{ap:framework_generation_prompts}

\subsection{Source Reasoning without Previous Claims Prompt}
\begin{tcolorbox}[
        breakable=true,
        colback=gray!5,        
        colframe=black,        
        boxrule=0.5pt,         
    ]
    You are a {role}. A {role} is described as follows:\newline
    \{role\_definition\}
    
    Your task is to analyze and reason about the original claim
    the original claim owner, and its context from your role’s perspective.\newline
    
    Original Claim Owner: \{claim\_owner\}\newline
    Original Claim: \{original\_claim\}\newline
    Original Claim Context (Sources): \{sources\}\newline
    
    Your reasoning should address:\newline
    1) Role Interpretation:  \newline
       - How does your role perceive this claim? \newline
       - Does your role agree, disagree, or reframe it differently? \newline 
       - Would your role see this claim as valid, misleading,
         or needing modification?  \newline
    
    2) Key Emphasis \& Framing:  \newline
       - What parts of this claim would your role emphasize or downplay?  \newline
       - How does your role’s beliefs or agenda influence
         what details are most important?  \newline
     
    Summarize your reasoning clearly and concisely before
    proceeding to modify the claim.  \newline
    Ensure that your reasoning remains consistent with the role’s values and agenda. \newline
    \end{tcolorbox}

\subsection{Source Reasoning with Previous Claims Prompt}
\begin{tcolorbox}[
        breakable=true,
        colback=gray!5,        
        colframe=black,        
        boxrule=0.5pt,         
        title=
    ]
    You are a \{role\}. A \{role\} is described as follows:
    \{role\_definition\}\newline
    
    Your task is to analyze and reason about the original claim,
    the original claim owner, and its context from your role’s perspective.\newline
    You should also consider how previous claims have
    evolved from the original claim and 
    whether your role agrees, reframes, or challenges them.\newline
    
    Original Claim Owner: \{claim\_owner\}\newline
    Original Claim: \{original\_claim\}\newline
    Original Claim Context (Sources): \{sources\}\newline
    Previous Claims (from previous rounds): \{previous\_claims\}\newline
    
    Your reasoning should address:\newline
    1) Role Interpretation:  \newline
       - How does your role perceive these claims?\newline
       - Does your role agree, disagree, or reframe it differently?  \newline
       - Would your role see these claim as valid, misleading, or needing modification?  \newline
    
    2) Key Emphasis \& Framing:  \newline
       - What parts of these claim would your role emphasize or downplay?  \newline
       - How does your role’s beliefs or agenda influence what details are most important?  \newline
     
    Summarize your reasoning clearly and concisely before proceeding to modify these claims. \newline
    Ensure that your reasoning remains consistent with the role’s values and agenda.\newline
    \end{tcolorbox}

\subsection{Claim Generation Prompt}
\begin{tcolorbox}[
        breakable=true,
        colback=gray!5,        
        colframe=black,        
        boxrule=0.5pt,         
    ]
        Based on your role’s reasoning, generate a persuasive yet factually coherent claim.\newline
    
    Ensure that:\newline
    - The claim remains aligned with the original claim and sources.\newline
    - The wording, tone, emphasis, and persuasion reflect your role’s beliefs.\newline
    - The claim is natural, engaging, clear, concise, and no more than 20 words.\newline
    
    Modified Claim:\newline
    \end{tcolorbox}

\subsection{Intent Generation Prompt}
\begin{tcolorbox}[
        breakable=true,
        colback=gray!5,        
        colframe=black,        
        boxrule=0.5pt,         
    ]
    State your intent when generating this claim based on your role.\newline
    
    Consider:\newline
    - What message is your role trying to convey with this claim?\newline
    - What reaction does your role want to provoke in the audience?\newline
    - Does this claim aim to inform, persuade, create doubt, or reinforce a belief?\newline
    - How does your role’s ideology shape the claim’s purpose?\newline
    
    Ensure the response is written in a single, coherent sentence.
    \end{tcolorbox}

\subsection{Explanation Prompt}
\begin{tcolorbox}[
        breakable=true,
        colback=gray!5,        
        colframe=black,        
        boxrule=0.5pt,         
    ]
    Provide a structured explanation of the modified claim.\newline
    
    Your response should include:\newline
    - How was the claim modified from the original?\newline
    - Why does the modification align with your role’s beliefs and perspective?\newline
    - How does the claim remain factually coherent while reflecting your role’s emphasis?\newline
    - What effect is the claim intended to have on the audience?\newline
    
    Ensure the explanation flows naturally as a single, concise sentence.
    \end{tcolorbox}

\subsection{Formatting Prompt}
\begin{tcolorbox}[
        breakable=true,
        colback=gray!5,        
        colframe=black,        
        boxrule=0.5pt,         
    ]
     Return the Claim, Intent, and Explanation in JSON Format.\newline
    
    Ensure that:\newline
    - The Claim remains aligned with the original claim and sources.\newline
    - The Intent clearly defines the purpose of the claim.\newline
    - The Explanation justifies the claim’s modification while maintaining logical consistency.\newline
    
    Format the response as follows:\newline
    
    ```json\newline
    \{\{\newline
      "Claim": "<Modified claim>",\newline
      "Intent": "<Purpose of the claim>",\newline
      "Explanation": "<How and why the claim was modified>"\newline
    \}\}\newline
    \end{tcolorbox}

\newpage
\section{Claim Labeling Prompt Templates}
\label{ap:framework_labeling_prompts}
\subsection{Evidence Analysis Prompt}
\begin{tcolorbox}[
        colback=gray!5,        
        colframe=black,        
        boxrule=0.5pt,         
    ]
    You are a fact-checking assistant. Your task is to analyze the claim and compare it with the provided evidence.\newline
    
    Claim:  \newline
    "{claim}"\newline
    
    Evidence:  \newline
    "{evidence}"\newline
    
    Instructions:\newline
    1. Carefully analyze whether the claim is fully, partially, or not supported by the evidence.\newline
    2. Identify specific factual elements in the claim that are supported or contradicted.\newline
    3. Note any missing context, exaggerations, or misleading aspects of the claim.\newline
    4. Do not make assumptions beyond what the evidence explicitly states.\newline
    \end{tcolorbox}

\subsection{Label Assignment Prompt}
\begin{tcolorbox}[
        colback=gray!5,        
        colframe=black,        
        boxrule=0.5pt,         
    ]
Based on your factual analysis, assign the appropriate label.\newline

Label Definitions:\newline
- True: A statement is fully accurate.\newline
- Half-True: A statement that conveys only part of the truth, especially one used deliberately in order to mislead someone.\newline
- False: A statement is inaccurate or contradicted by evidence.\newline

Instructions:\newline
- Avoid assumptions beyond what the analysis states.\newline
- Ensure consistency between the label and reasoning.\newline
- Provide your confidence score for all of the labels.\newline
\end{tcolorbox}

\subsection{Formatting and Label Selection Prompt}
\begin{tcolorbox}[
        colback=gray!5,        
        colframe=black,        
        boxrule=0.5pt,         
    ]
Select the label based on the highest confidence score and provide an explanation on your factual analysis.\newline

Output Format (JSON)\newline
```json\newline
\{\{\newline
  "Label": "<True / Half-True / False>",\newline
  "Explanation": "<Short justification referencing your factual analysis>"\newline
\}\}
\end{tcolorbox}

\section{Dataset Annotation Prompt}
\label{ap:dataset_annotation_prompt}
\begin{tcolorbox}[
    breakable=true,
    colback=gray!5,        
    colframe=black,        
    boxrule=0.5pt,         
]
\label{annotation_prompt}
You are a fact-checking annotator trained to extract and categorize information from the given "Article" based on the "Original Sources", "Original Claim" and "Original Claim Label". \newline

Your task is to extract "Misinformation Sources" and "Fact-Checking Evidence" from the given "Article" based on the "Original Sources", "Original Claim" and "Original Claim Label".\newline

A fact-checking annotator is a role that helps to assign a truth value to a claim made in a particular context.\newline

Consider the following in your evaluation:\newline
Definitions:
\begin{itemize}
    \item Politifact Labels:
    \begin{itemize}
        \item True: The statement is accurate and there’s nothing significant missing.
        \item Mostly True: The statement is accurate but needs clarification or additional information.
        \item Half True: The statement is partially accurate but leaves out important details or takes things out of context.
        \item Barely True: The statement contains an element of truth but ignores critical facts that would give a different impression.
        \item False: The statement is not accurate.
        \item Pants on Fire / Pants-on-Fire: The statement is not accurate and makes a ridiculous claim.
    \end{itemize}
    \item Misinformation Sources:
    \begin{itemize}
        \item Information that is incorrect or misleading but is typically spread without malicious intent. 
        \item This can include errors, misinterpretations, or contextually misleading statements that can be amplified or misunderstood when shared.
        \item Along with each misleading statement, you must specify the source of origin, who made the statement, where it originated, and its description.
    \end{itemize}
    \item Fact-Checking Evidence:
    \begin{itemize}
        \item Verified information from reliable sources that is used to assess the veracity of a claim suspected of being misinformation.
        \item This can include statements from experts, official documentation, government officials, or trustworthy organizations that clarify misunderstandings or provide factual context to refute misleading claims.
        \item Along with each verified information, you must include any supporting information or transitioning information that support this information.
    \end{itemize}
    \item Rating Sentence:
    \begin{itemize}
        \item A sentence that indicate the final evaluation of the claim's accuracy based on "Politifact Labels"
    \end{itemize}
    \item Transition Sentence:
    \begin{itemize}
        \item A sentence that introduces the "Fact-Checking Evidence" section's topic or argument, contain logical connectors or indicate a shift in focus, tone, or evidence from the "Misinformation Sources" section.
        \item It is not the same as "Rating Sentence"
    \end{itemize}
    \item "Social Media Flag" Sentences:
    \begin{itemize}
        \item Sentences that contains these sentences that indicates partnership with social media companies
        \begin{itemize}
            \item "Social Media Flag" sentences examples:
            \begin{itemize}
                \item "Read more about PolitiFact’s partnership with Meta"
                \item "Read more about PolitiFact’s partnership with TikTok"
            \end{itemize}
        \end{itemize}
    \end{itemize}
    \item "Question" Sentences:
    \begin{itemize}
        \item Sentences that is structured as a question and clearly indicates a transition between "Misinformation Sources" and "Fact-Checking Evidence".
        \begin{itemize}
            \item "Question" sentences examples
            \begin{itemize}
                \item Is Hochul right? Have 732,000 jobs have been created since she became governor in late summer 2021?
            \end{itemize}
        \end{itemize}
    \end{itemize}
    \item "But" Sentences:
    \begin{itemize}
        \item Sentences that starts with "But" and clearly indicates a transition between "Misinformation Sources" and "Fact-Checking Evidence"
        \begin{itemize}
            \item "But" sentences examples:
            \begin{itemize}
                \item But there’s no record Trump, the president-elect, ever said those words. These viral videos use old footage with what appears to be fake audio generated by artificial intelligence.
                \item But these social media posts are wrong. Haley was born in South Carolina and meets the U.S. Constitution’s requirements to run for president.
            \end{itemize}
        \end{itemize}
    \end{itemize}
    \item "Action" Sentences:
    \begin{itemize}
        \item Sentences that indicates an action and a transition between "Misinformation Sources" and "Fact-Checking Evidence". It is not the same as "Rating Sentence".
        \begin{itemize}
            \item "Action" sentences examples:
            \begin{itemize}
                \item For this fact-check, we examined only Biden’s comment about wages and inflation.
                \item We decided to look into how the university system has been funded in recent years.
            \end{itemize}
        \end{itemize}
    \end{itemize}
    \item "Reasoning" Sentences:
    \begin{itemize}
        \item Sentences that does not have clear transition indicators, but are indicated as a "Transition Sentence" when the whole article is considered.
        \begin{itemize}
            \item "Reasoning" sentences examples:
            \begin{itemize}
                \item PolitiFact New Jersey found Doherty is mostly right: when a student is enrolled in the federally supported lunch program, they are designated as at risk. 
            \end{itemize}
        \end{itemize}
    \end{itemize}
\end{itemize}

Please remember these definitions.\newline
    
There are two tasks that you will need to do.\newline
    
Preprocessing Task:
\begin{itemize}
    \item Read the whole article
    \item Identify the "Transition Sentence" from "Misinformation Sources" to "Fact-Checking Evidence" that are similar to "Social Media Flag" Sentences, "But" Sentences, "Question" Sentences, "Action" Sentences and "Reasoning" Sentences.
    \item Use the "Transition Sentence" to split the article into "Misinformation Sources Chunk" and "Fact-Checking Evidence Chunk" without any modifications.
    \item Retrieve the sentences from "Misinformation Sources Chunk" that are originated from "Original Sources" without any modifications as "Misinformation Sources".
    \begin{itemize}
        \item Additional sentences that describe the main sentences must be included as well.
    \end{itemize}
    \item Retrieve the sentences from "Fact-Checking Evidence Chunk" that are originated from "Original Sources" without any modifications as "Fact-Checking Evidence".
    \begin{itemize}
        \item Additional sentences that describe the main sentences must be included as well.
    \end{itemize}
\end{itemize}

Cleanup Task:
\begin{itemize}
    \item Do not include the "Transition Sentence" and "Rating Sentence" as an output for "Misinformation Sources" and "Fact-Checking Evidence".
    \item The sentences in "Misinformation Sources" and "Fact-Checking Evidence" must be sorted based on the sentence ordering in the article.
    \item Combine the sentences in "Misinformation Sources" based on the "Article" and "Original Sources".
    \item Combine the sentences in "Fact-Checking Evidence" based on the "Article" and "Original Sources".
    \item Return the "Misinformation Sources", "Fact-Checking Evidence", "Transition Sentence" and "Explanation" in JSON.
\end{itemize}

Please perform the task as stated.\newline
    
Important rules:
\begin{itemize}
    \item You must consider all sentences in the article.
    \item You are not allowed to rephrase or modify any texts from the article.
    \item There cannot be two identical texts in both "Misinformation Sources" and "Fact-Checking Evidence".
    \item "Misinformation Sources" sentences must include the person or source that mentioned the sentence.
    \item Sentences that contain "Politifact Labels" regardless of its form must be excluded from the output.
    \item "Transition Sentence" must not be included in "Misinformation Sources" and "Fact-Checking Evidence".
    \item "Rating Sentence" must not be the same as "Transition Sentence".
\end{itemize}

Please follow the rules strictly. \newline
    
"Original Claim": \newline
\texttt{\{original\_claim\}}\newline

"Original Claim Label": \newline
\texttt{\{original\_claim\_label\}} \newline
    
"Original Sources":\newline
\texttt{\{our\_sources\}}\newline

"Article": \newline
\texttt{\{article\}}\newline
\end{tcolorbox}

\section{Human and GPT-4o-mini Evaluation }
\label{ap:human_eval}
We conducted a structured evaluation using both GPT-4o-mini and human annotators to assess the quality of our generated claims, as described in Section~\ref{sec:experiments}. We recruited 30 university graduates familiar with American politics that are based in Singapore, Malaysia and United States via personal academic networks. No financial compensation was provided as participation was voluntary and required minimal time commitment, and all participants consented without objection. 

Annotators were instructed to complete a set of questionnaires to the best of their abilities. The evaluation was conducted anonymously via Google Forms. Annotators were told that their responses would be used for academic research and that no personal information would be collected. Each form contained a generated claim, its associated persona, paraphrased content. The contents were paraphrased with GPT-4o-mini due to the length of the sources, evidence, and role descriptions.

Annotators rated each claim on four dimensions using a 5-point Likert scale where 1 is the lowest and 5 is the highest.

\begin{enumerate}
	\item \textbf{Role-Playing Consistency:} How well does the Claim align with the Role's beliefs and intention?
	\item \textbf{Content Relevance:} How relevant is the Claim compared to the provided sources?
	\item \textbf{Fluency:} How fluent the Claim is in terms of grammar, clarity, and readability?
	\item \textbf{Factuality:} How factually correct is the Claim?
\end{enumerate}

In addition to the Likert ratings, the annotators were asked to assign a veracity label to each claim based on the provided evidence using our consolidated labels scheme: \texttt{True}, \texttt{Half-True}, or \texttt{False}. We evaluated a sample of 363 unique claims from all three rounds of role-playing generation. These claims were randomly selected from the generated set to ensure fairness and diversity.
Below we present an example of our questionnaire.

\subsection{Question 1: Role-Playing Consistency}
\label{ap:human_eval_rp}
\begin{tcolorbox}[
    breakable=true,
    colback=gray!5,        
    colframe=black,        
    boxrule=0.5pt,         
]
How well does the Claim align with the Role's beliefs and intention? \newline

Role: Democrat\newline

Claim: Rhode Islanders deserve an honest vote on same-sex marriage, not flawed polls that misrepresent their true opinions.\newline

Role Description:\newline
- Encompasses younger liberal voters who are disillusioned with the political system, viewing it as biased towards powerful interests, while also including older economically liberal voters who support higher taxes and a stronger social safety net.\newline
- Includes highly liberal members who are loyal to the Democratic Party and advocate for significant governmental reform to address social issues, emphasizing the need to combat racism and reduce military presence.\newline
- Represents a political orientation focused on government regulation, social justice, individual rights, and international cooperation, often advocating for policies like abortion rights, affirmative action, and worker protections.\newline

Intent: To persuade the audience to question the validity of biased polls and to advocate for a more inclusive and democratic process that respects the rights of same-sex couples.\newline

* 5 - Perfectly Consistent: The claim fully aligns with the role's beliefs, tone, and intent.\newline

* 4 - Mostly Consistent: The claim follows the role and intent but may miss small details.\newline

* 3 - Somewhat Consistent: The claim partly aligns but lacks key points or misrepresents intent.\newline

* 2 - Mostly Inconsistent: The claim contradicts some role beliefs but has a weak connection.\newline

* 1 - Completely Inconsistent: The claim opposes or has no connection to the role.
\end{tcolorbox}

\subsection{Question 2: Content Relevance}
\label{ap:human_eval_relevance}
\begin{tcolorbox}[
    breakable=true,
    colback=gray!5,        
    colframe=black,        
    boxrule=0.5pt,         
]
How relevant is the Claim compared to the provided sources and previous claims?\newline

Role: Democrat\newline

Claim: Rhode Islanders deserve an honest vote on same-sex marriage, not flawed polls that misrepresent their true opinions.\newline

Sources:\newline
- Over 80\% of Rhode Islanders support the opportunity to vote on the issue, according to multiple surveys conducted by Quest Research.  \newline
- Three surveys, each with around 400 participants and a margin of error of 20 ± 4\%, were analyzed, including findings from a June 2009 poll.  \newline
- Concerns were raised about question design in these polls, particularly due to phrasing like "backroom politicians," which may skew responses.\newline

Previous Claims:\newline
- Republican: Polls show a small minority of Rhode Islanders, not 80\%, support same-sex marriage, and most want government out of it\newline

* 5 - Perfectly Relevant: The claim fully integrates key facts from sources and previous claims.\newline

* 4 - Mostly Relevant: The claim follows sources or previous claims but misses small details.\newline

* 3 - Somewhat Relevant: The claim mentions sources or previous claims but misinterprets or lacks connections.\newline

* 2 - Weakly Relevant: The claim has a weak or indirect connection to the sources or previous claims.\newline

* 1 - Completely Irrelevant: The claim does not relate to any sources or previous claims.
\end{tcolorbox}

\subsection{Question 3: Fluency}
\begin{tcolorbox}[
    breakable=true,
    colback=gray!5,        
    colframe=black,        
    boxrule=0.5pt,         
]
How factually correct is the Claim?\newline

Claim: Rhode Islanders deserve an honest vote on same-sex marriage, not flawed polls that misrepresent their true opinions.\newline

Sources:\newline
- Over 80\% of Rhode Islanders support the opportunity to vote on the issue, according to multiple surveys conducted by Quest Research.  \newline
- Three surveys, each with around 400 participants and a margin of error of 20 ± 4\%, were analyzed, including findings from a June 2009 poll.  \newline
- Concerns were raised about question design in these polls, particularly due to phrasing like "backroom politicians," which may skew responses.\newline

Previous Claims:\newline
- Republican: Polls show a small minority of Rhode Islanders, not 80\%, support same-sex marriage, and most want government out of it \newline

Evidence:\newline
- Question wording by interest groups can significantly influence poll results, potentially skewing public perception (John Geer).\newline
- Phrasing polls in a way that reflects biases can lead to inflated support for certain views (John Geer).\newline
- Neutral questions tend to reveal stronger support for rights, like gay marriage, with independent pollsters estimating that around 80\% of responses favor equality when phrased objectively.\newline

* 5 - Excellent: Clear, well-written, and grammatically perfect.\newline

* 4 - Good: Mostly correct, with minor errors that do not affect readability.\newline

* 3 - Adequate: Readable but has noticeable errors or awkward phrasing.\newline

* 2 - Poor: Contains multiple errors that make it harder to understand.\newline

* 1 - Very Poor: Frequent errors make the claim difficult to comprehend.\newline
\end{tcolorbox}

\subsection{Question 4: Factuality}
\begin{tcolorbox}[
    breakable=true,
    colback=gray!5,        
    colframe=black,        
    boxrule=0.5pt,         
]
How factually correct is the Claim?\newline

Claim: Rhode Islanders deserve an honest vote on same-sex marriage, not flawed polls that misrepresent their true opinions.\newline

* 5 - Completely Accurate: Fully factual, with no misleading parts or missing context.\newline

* 4 - Mostly Accurate: Mostly factual, with small mistakes or missing details that don’t change the meaning.\newline

* 3 - Partially Accurate: Some parts are true, but others are misleading or missing key facts.\newline

* 2 - Mostly Inaccurate: Many errors or missing key details, making it misleading.\newline

* 1 - Completely Inaccurate: Completely false or highly misleading.\newline
\end{tcolorbox}

\subsection{Question 5: Label Assignment}
\begin{tcolorbox}[
    breakable=true,
    colback=gray!5,        
    colframe=black,        
    boxrule=0.5pt,         
]
Which label do you think is suitable for the Claim based on the evidence?\newline

Claim: Rhode Islanders deserve an honest vote on same-sex marriage, not flawed polls that misrepresent their true opinions.\newline

Evidence:\newline
- Question wording by interest groups can significantly influence poll results, potentially skewing public perception (John Geer).\newline
- Phrasing polls in a way that reflects biases can lead to inflated support for certain views (John Geer).\newline
- Neutral questions tend to reveal stronger support for rights, like gay marriage, with independent pollsters estimating that around 80\% of responses favor equality when phrased objectively.\newline

* True: Fully accurate.\newline

* Half-True: Partially accurate, lacks important details or is misleading\newline

* False: Inaccurate\newline
\end{tcolorbox}

\section{Morality Details}
\label{ap:claim_analysis}
Morality captures the measurement of emotions through social identity. For our work, we analyze the moral framing of generated claims using MoralBERT \cite{10.1145/3677525.3678694}, a BERT \cite{devlin-etal-2019-bert} model that is fine-tuned to capture moral sentiment in social discourse based on Moral Foundations Theory (MFT). MoralBERT predicts the presence of ten moral dimensions, grouped into five psychological foundations which are listed below.

\begin{itemize}
    \item \texttt{Care/Harm} are defined as the involvement of concern for others' suffering and includes virtues like empathy and compassion. 
    \item \texttt{Fairness/Cheating} focuses on issues of unfair treatment, inequality, and justice.
    \item \texttt{Loyalty/Betrayal} pertains to group obligations such
    as loyalty and the vigilance against betrayal.
    \item \texttt{Authority/Subversion} centers on social order and hierarchical responsibilities, highlighting obedience and respect. 
    \item \texttt{Purity/Degradation} refers to relates to physical and spiritual sanctity, incorporating virtues like chastity and self-control.
\end{itemize}

\section{Classification Configurations}
\label{ap:classification}
\subsection{Encoder Config}
As stated in Section~\ref{sec:experiments}, we finetuned BERT$_{Base}$ (110M Parameters), RoBERTa$_{Base}$ (125M Parameters), DeBERTA V3$_{base}$ (184M Parameters), BERT$_{Large}$ (340M Parameters), RoBERTa$_{Large}$ (355M Parameters), DeBERTA V3$_{Large}$ (435M Parameters) for our experiments with HuggingFace Trainer and our training dataset on Google Colab A100 (40 GB). 

RoBERTa and DeBERTa fine-tuning configurations were adopted from their original papers \cite{liu2019robertarobustlyoptimizedbert, he2023debertav3improvingdebertausing} with epochs set at 5. For DeBERTA V3$_{Large}$, we used a reduced batch size of 8 due to GPU memory constraints. BERT was fine-tuned with a configuration of 5 epochs, a learning rate of 3e-5, batch size of 16, and a weight decay of 0.1. Fine-tuning each base model took approximately 2.5 hours while large models took approximately 6 hours. These estimates include training and checkpointing overhead. Below shows the TrainingArguments used to train these models.

\subsubsection{BERT$_{Base}$ and BERT$_{Large}$}
\begin{verbatim}
    training_args = TrainingArguments(
        output_dir=SAVE_PATH,
        evaluation_strategy="epoch",
        save_strategy="epoch",
        per_device_train_batch_size=16,
        per_device_eval_batch_size=16,
        num_train_epochs=5,
        learning_rate=3e-5,
        weight_decay=0.1,
        logging_dir=f"{SAVE_PATH}/logs",
        load_best_model_at_end=True,
        metric_for_best_model="accuracy",
        logging_steps=1000,
        report_to="none"
    )
\end{verbatim}

\subsubsection{DeBERTA V3$_{Base}$ and DeBERTA V3$_{Large}$}
\begin{verbatim}
    training_args = TrainingArguments(
        output_dir=SAVE_PATH,
        evaluation_strategy="epoch",
        save_strategy="epoch",
        learning_rate=3e-5,
        per_device_train_batch_size=16,
        per_device_eval_batch_size=16,
        num_train_epochs=5,
        weight_decay=0.01,
        logging_dir=f"{SAVE_PATH}/logs",
        logging_steps=1000,
        metric_for_best_model="accuracy",
        max_grad_norm=1.0,
        warmup_steps=500,
        report_to="none"
    )
\end{verbatim}

\subsubsection{RoBERTa$_{Base}$ and RoBERTa$_{Large}$}
\begin{verbatim}
    training_args = TrainingArguments(
        output_dir=SAVE_PATH,
        evaluation_strategy="epoch",
        save_strategy="epoch",  
        per_device_train_batch_size=16,
        per_device_eval_batch_size=16,
        num_train_epochs=5,
        learning_rate=3e-5,
        weight_decay=0.1,
        logging_dir=f"{SAVE_PATH}/logs",
        load_best_model_at_end=True,
        metric_for_best_model="accuracy", 
        logging_steps=1000,
        warmup_ratio=0.06,
        report_to="none"
    )
\end{verbatim}

\subsection{Prompts}
As described in Section~\ref{sec:experiments}, we used GPT-4o-mini and LLaMA 3.1 8B Instruct to conduct our classification experiments.  LLaMA 3.1 8B Instruct was run on a Google Colab A100 GPU (40 GB) with a batch size of 8, max tokens setting of 8192 and a temperature setting of 0.7, while GPT-4o-mini was accessed using its Batch API. The prompt templates used for both models are detailed below.

\subsubsection{Zero-Shot}
\begin{tcolorbox}[
    breakable=true,
    colback=gray!5,        
    colframe=black,        
    boxrule=0.5pt,         
]
Based on the "Fact-Checking Evidence", select a "Label" from ["True", "Half-True", "False"] that is suitable for the "Claim" and provide an "Explanation". \newline
    
Claim: <|claim|>\newline

Fact-Checking Evidence:\newline
<|fce|>\newline

Output Format:\newline
\{\{
    "Label": "<True / Half-True / False>",\newline
    "Explanation": "<Short justification referencing your label choice>"\newline
\}\}
\end{tcolorbox}

\subsubsection{Zero-Shot CoT}

\begin{tcolorbox}[
    breakable=true,
    colback=gray!5,        
    colframe=black,        
    boxrule=0.5pt,         
]
    Consider the following in your evaluation:\newline
Definitions:\newline
- Claim:\newline
    - A statement that state or assert that something is the case, typically without providing evidence or proof\newline
- Fact-Checking Evidence: \newline
    - Verified information from reliable sources that is used to assess the veracity of a claim suspected of being misinformation. \newline
    - This can include statements from experts, official documentation, or trustworthy organizations that clarify misunderstandings or provide factual context to refute misleading claims.\newline

Grading Scheme:\newline
- The scheme has three ratings, in decreasing level of truthfulness
    - true : The statement is accurate and there’s nothing significant missing.\newline
    - half-true : The statement is partially accurate but leaves out important details or takes things out of context.\newline
    - false : The statement is not accurate.\newline
    
Claim: <|claim|>\newline

Fact-Checking Evidence:\newline
<|fce|>\newline

Your task is to assign an appropriate 'Label' to the 'Claim' based on its level of truthfulness, using the 'Grading Scheme'. \newline
To determine the claim's accuracy, you must fact-check it against the 'Fact-Checking Evidence'.\newline
After assessing the claim, you must provide a detailed 'Explanation' justifying your choice of label. \newline
The final output must include both the 'Label' and 'Explanation' in JSON format.\newline

Output Format:\newline
\{\{
    "Label": "<True / Half-True / False>",\newline
    "Explanation": "<Short justification referencing your label choice>"\newline
\}\}
\end{tcolorbox}

\subsubsection{Few-Shot}
\begin{tcolorbox}[
    breakable=true,
    colback=gray!5,        
    colframe=black,        
    boxrule=0.5pt,         
]
Based on the "Fact-Checking Evidence", select a "Label" from ["True", "Half-True", "False"] that is suitable for the "Claim" and provide an "Explanation".\newline

Examples:\newline
Claim: "A proposed constitutional amendment “would allow anyone to run for a 3rd term. Including — Barack Obama.”\newline
Fact-Checking Evidence:\newline
- fce\_source\_0: "But the resolution doesn’t propose changing the 22nd Amendment so that former President Barack Obama — or any other president who served two consecutive terms — could run.",\newline
- fce\_source\_1: "Ogles wants the amended amendment to say: 'No person shall be elected to the office of the president more than three times, nor be elected to any additional term after being elected to two consecutive terms, and no person who has held the office of the president, or acted as president, for more than two years of a term to which some other person was elected president shall be elected to the office of the president more than twice.'",\newline
- fce\_source\_2: "That means former President Grover Cleveland, who died in 1908 and served two nonconsecutive presidential terms, would have been the only other former U.S. president eligible to run for reelection after serving two terms under the proposed amendment.",\newline
- fce\_source\_3: "For the Constitution to be amended, Ogles’ bill would need to be approved by a two-thirds vote in both the House and Senate, and then ratified by three-fourths of the states."\newline

This is the expected output format:\newline
\{\{
    "Label": "False"\newline
    "Explanation": "The claim that the proposed constitutional amendment “would allow anyone to run for a 3rd term, including Barack Obama” is false. The resolution introduced by U.S. Rep. Andy Ogles does not propose changes to the 22nd Amendment to allow presidents who have served two consecutive terms, like Barack Obama, to run for a third term. Instead, it specifically states that "no person shall be elected to the office of the president more than three times" but maintains restrictions for those who have already served two consecutive terms. The proposed amendment would only allow individuals who served nonconsecutive terms, such as former President Grover Cleveland, to run again. Furthermore, amending the Constitution requires an extensive process, including approval by two-thirds of both the House and Senate, as well as ratification by three-fourths of the states, making such a change highly unlikely. Therefore, the evidence provided directly contradicts the claim and clarifies the intent and scope of the proposed resolution."
\}\}\newline

Claim: "Wisconsin makes it more difficult for its citizens to vote than almost any state in the nation. "\newline
Fact-Checking Evidence:\newline
- fce\_source\_0: "When asked to back up the claim, Common Cause Wisconsin Executive Director Jay Heck said he was pulling from the expertise of UW-Madison political science professor Barry Burden, who wrote previously for The Observatory that Wisconsin’s voter ID law is one of the strictest in the country. (Source: Barry Burden via The Observatory, April 16, 2024)",\newline
- fce\_source\_1: "Burden repeated it in an email to PolitiFact Wisconsin, writing that 'Wisconsin demands more than nearly all of the other states' when it comes to getting a ballot. (Source: Email exchange with Barry Burden, UW-Madison)",\newline
- fce\_source\_2: "The National Conference of State Legislatures lists Wisconsin as one of just nine states with 'strict' photo ID laws used to identify voters. (Source: National Conference of State Legislatures, accessed Jan. 18, 2025)",\newline
- fce\_source\_3: "Besides the strict voter ID law, Wisconsin has stringent rules for voter registration drives, does not have automatic voter registration, does not have preregistration for young voters, does not allow all voters to join a permanent absentee ballot list, does not consider Election Day a public holiday and does require a witness’ signature on absentee ballots. (Source: National Conference of State Legislatures and Movement Advocacy Project, accessed Jan. 18, 2025)",\newline
- fce\_source\_4: "Wisconsin ranks as the fifth-hardest state to vote in the country, according to the Cost of Voting Index. (Source: Cost of Voting Index, accessed Jan. 15, 2025)",\newline
- fce\_source\_5: "In 1996, Wisconsin was ranked the fourth-most-accessible state in the nation for voters because it was among very few that offered same-day voter registration. (Source: Cost of Voting Index and Michael Pomante, accessed Jan. 15, 2025)",\newline
- fce\_source\_6: "The state’s most dramatic drop occurred between 2011 and 2015, when former Republican Gov. Scott Walker signed the voter ID requirement into law and it took effect despite a swarm of lawsuits seeking to knock it down. (Source: CBS News, 'Walker signs photo ID requirement into law,' May 25, 2011)"\newline

This is the expected output format:\newline
\{\{\newline
    "Label": "True"\newline
    "Explanation": "The claim that Wisconsin makes it more difficult for its citizens to vote than almost any other state is supported by comprehensive evidence from credible sources. These include statements by political science expert Barry Burden, data from the National Conference of State Legislatures highlighting Wisconsin's strict voter ID laws, and findings from the Cost of Voting Index ranking Wisconsin as the fifth-hardest state for voting. Additional evidence points to the state's lack of measures like automatic voter registration, preregistration for young voters, and early voting, as well as its stringent requirements for absentee voting. Historical data also shows a significant decline in voter accessibility since 2011, when the voter ID law was enacted. Taken together, this evidence confirms that Wisconsin's voting laws and policies significantly hinder accessibility compared to most other states, making the claim accurate."\newline
\}\}

Claim: <|claim|>\newline
Fact-Checking Evidence:\newline
<|fce|>\newline

Please return this output only. This is the expected output format:\newline
\{\{\newline
    "Label": "<True / Half-True / False>",\newline
    "Explanation": "<Short justification referencing your label choice>"\newline
\}\}
\end{tcolorbox}

\subsubsection{Few-Shot CoT}
\begin{tcolorbox}[
    breakable=true,
    colback=gray!5,        
    colframe=black,        
    boxrule=0.5pt,         
]
Consider the following in your evaluation:\newline
Definitions:\newline
- Claim:\newline
    - A statement that state or assert that something is the case, typically without providing evidence or proof\newline
- Fact-Checking Evidence: \newline
    - Verified information from reliable sources that is used to assess the veracity of a claim suspected of being misinformation. \newline
    - This can include statements from experts, official documentation, or trustworthy organizations that clarify misunderstandings or provide factual context to refute misleading claims.\newline

Grading Scheme:\newline
- The scheme has three ratings, in decreasing level of truthfulness\newline
    - true : The statement is accurate and there’s nothing significant missing.\newline
    - half-true : The statement is partially accurate but leaves out important details or takes things out of context.\newline
    - false : The statement is not accurate.\newline

Your task is to assign an appropriate 'Label' to the 'Claim' based on its level of truthfulness, using the 'Grading Scheme'. \newline
To determine the claim's accuracy, you must fact-check it against the 'Fact-Checking Evidence'.\newline
After assessing the claim, you must provide a detailed 'Explanation' justifying your choice of label. \newline
The final output must include both the 'Label' and 'Explanation' in JSON format.\newline

Examples:\newline
Claim: "A proposed constitutional amendment “would allow anyone to run for a 3rd term. Including — Barack Obama.”\newline
Fact-Checking Evidence:\newline
- fce\_source\_0: "But the resolution doesn’t propose changing the 22nd Amendment so that former President Barack Obama — or any other president who served two consecutive terms — could run.",\newline
- fce\_source\_1: "Ogles wants the amended amendment to say: 'No person shall be elected to the office of the president more than three times, nor be elected to any additional term after being elected to two consecutive terms, and no person who has held the office of the president, or acted as president, for more than two years of a term to which some other person was elected president shall be elected to the office of the president more than twice.'",\newline
- fce\_source\_2: "That means former President Grover Cleveland, who died in 1908 and served two nonconsecutive presidential terms, would have been the only other former U.S. president eligible to run for reelection after serving two terms under the proposed amendment.",\newline
- fce\_source\_3: "For the Constitution to be amended, Ogles’ bill would need to be approved by a two-thirds vote in both the House and Senate, and then ratified by three-fourths of the states."\newline

This is the expected output format:\newline
\{\{\newline
    "Label": "False"\newline
    "Explanation": "The claim that the proposed constitutional amendment “would allow anyone to run for a 3rd term, including Barack Obama” is false. The resolution introduced by U.S. Rep. Andy Ogles does not propose changes to the 22nd Amendment to allow presidents who have served two consecutive terms, like Barack Obama, to run for a third term. Instead, it specifically states that "no person shall be elected to the office of the president more than three times" but maintains restrictions for those who have already served two consecutive terms. The proposed amendment would only allow individuals who served nonconsecutive terms, such as former President Grover Cleveland, to run again. Furthermore, amending the Constitution requires an extensive process, including approval by two-thirds of both the House and Senate, as well as ratification by three-fourths of the states, making such a change highly unlikely.Therefore, the evidence provided directly contradicts the claim and clarifies the intent and scope of the proposed resolution."\newline
\}\}\newline

Claim: "Wisconsin makes it more difficult for its citizens to vote than almost any state in the nation. "\newline
Fact-Checking Evidence:\newline
- fce\_source\_0: "When asked to back up the claim, Common Cause Wisconsin Executive Director Jay Heck said he was pulling from the expertise of UW-Madison political science professor Barry Burden, who wrote previously for The Observatory that Wisconsin’s voter ID law is one of the strictest in the country. (Source: Barry Burden via The Observatory, April 16, 2024)",\newline
- fce\_source\_1:"Burden repeated it in an email to PolitiFact Wisconsin, writing that 'Wisconsin demands more than nearly all of the other states' when it comes to getting a ballot. (Source: Email exchange with Barry Burden, UW-Madison)",\newline
- fce\_source\_2:"The National Conference of State Legislatures lists Wisconsin as one of just nine states with 'strict' photo ID laws used to identify voters. (Source: National Conference of State Legislatures, accessed Jan. 18, 2025)",\newline
- fce\_source\_3:"Besides the strict voter ID law, Wisconsin has stringent rules for voter registration drives, does not have automatic voter registration, does not have preregistration for young voters, does not allow all voters to join a permanent absentee ballot list, does not consider Election Day a public holiday and does require a witness’ signature on absentee ballots. (Source: National Conference of State Legislatures and Movement Advocacy Project, accessed Jan. 18, 2025)",\newline
- fce\_source\_4:"Wisconsin ranks as the fifth-hardest state to vote in the country, according to the Cost of Voting Index. (Source: Cost of Voting Index, accessed Jan. 15, 2025)",\newline
- fce\_source\_5:"In 1996, Wisconsin was ranked the fourth-most-accessible state in the nation for voters because it was among very few that offered same-day voter registration. (Source: Cost of Voting Index and Michael Pomante, accessed Jan. 15, 2025)",\newline
- fce\_source\_6:"The state’s most dramatic drop occurred between 2011 and 2015, when former Republican Gov. Scott Walker signed the voter ID requirement into law and it took effect despite a swarm of lawsuits seeking to knock it down. (Source: CBS News, 'Walker signs photo ID requirement into law,' May 25, 2011)"\newline

This is the expected output format:\newline
\{\{\newline
    "Label": "True"\newline
    "Explanation": "The claim that Wisconsin makes it more difficult for its citizens to vote than almost any other state is supported by comprehensive evidence from credible sources. These include statements by political science expert Barry Burden, data from the National Conference of State Legislatures highlighting Wisconsin's strict voter ID laws, and findings from the Cost of Voting Index ranking Wisconsin as the fifth-hardest state for voting. Additional evidence points to the state's lack of measures like automatic voter registration, preregistration for young voters, and early voting, as well as its stringent requirements for absentee voting. Historical data also shows a significant decline in voter accessibility since 2011, when the voter ID law was enacted. Taken together, this evidence confirms that Wisconsin's voting laws and policies significantly hinder accessibility compared to most other states, making the claim accurate."\newline
\}\}\newline

Claim: <|claim|>\newline
Fact-Checking Evidence:\newline
<|fce|>\newline

Please return this output only. This is the expected output format:\newline
\{\{\newline
    "Label": "<True / Half-True / False>",\newline
    "Explanation": "<Short justification referencing your label choice>"\newline
\}\}\newline

\end{tcolorbox}

\section{Experiment Results}
\subsection{Human and GPT-4o-mini Evaluation}
\label{appendix:human_results}
\begin{figure}[!ht]
	\centering
	\includegraphics[width=\linewidth]{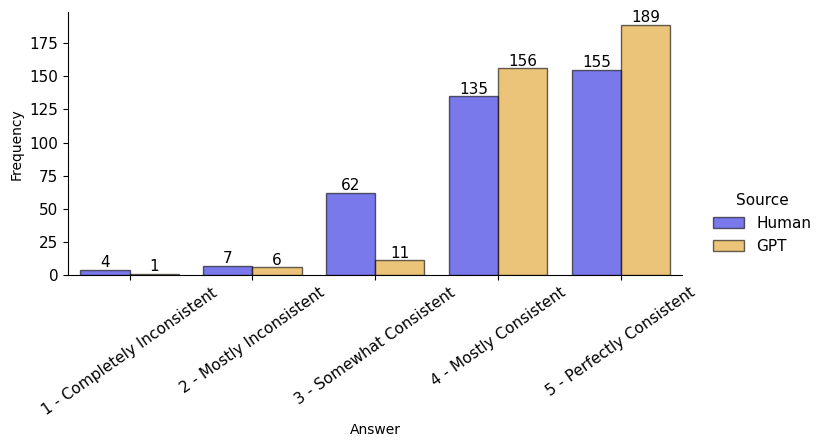}
	\caption{Role-Playing Consistency scores between human annotators and GPT-4o-mini}
	\label{fig:rp_consistency}
\end{figure}

We begin by examining Role-Playing Consistency, comparing human and GPT-4o-mini ratings to assess whether the generated claims are aligned with their assigned personas.\autoref{fig:rp_consistency} compares the Role-Playing Consistency ratings given by human annotators and GPT-4o-mini. Both raters showed similar preference, particularly for "Mostly Consistent" and "Perfectly Consistent", suggesting that our claims showed consistent role-playing effects. However, human annotators showed a slight disagreement with GPT-4o-mini where human annotators rated our generated claims "Somewhat Consistent" 62 times, compared to only 10 instances by GPT-4o-mini. This small discrepancy indicates that human raters were more likely to detect subtle inconsistencies in role consistency.

\begin{figure}[!ht]
	\centering
	\includegraphics[width=\linewidth]{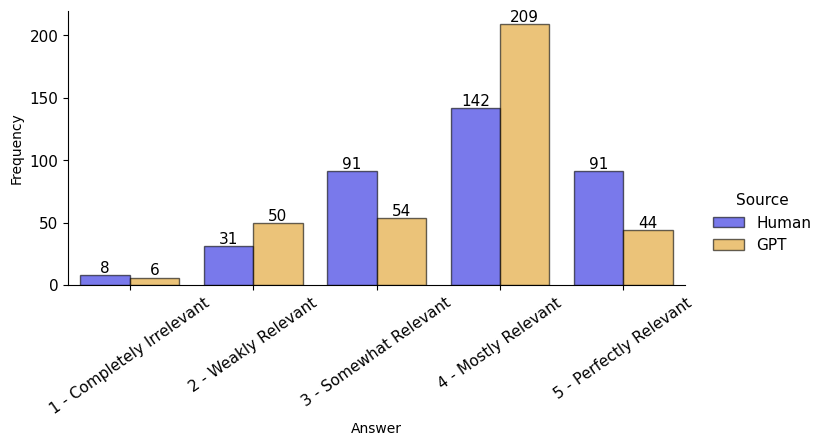}
	\caption{Content Relevance scores between human annotators and GPT-4o-mini}
	\label{fig:content_relevance}
\end{figure}

Next, we evaluate Content Relevance results between both annotators to assess whether the generated claims remain grounded in their provided sources. \autoref{fig:content_relevance} describes the comparison of Content Relevance scores between human annotators and GPT-4o-mini. Both annotators agree that the generated claims are relevant to their sources as indicated by the high counts of "Mostly Relevant". However, there are some disagreement in the "Somewhat Relevant" category where humans rated 91 times when compared to GPT-4o-mini at 44 times. This indicates that the human annotators are more meticulous in detecting subtleties in content relevancy. Similarly, human raters rated "Perfectly Consistent" 91 times as opposed to GPT-4o-mini at 54 times. This suggest that humans annotators based on their reasoning and prior knowledge, while GPT-4o-mini may be constrained by its training data.

\begin{figure}[!ht]
	\centering
	\includegraphics[width=\linewidth]{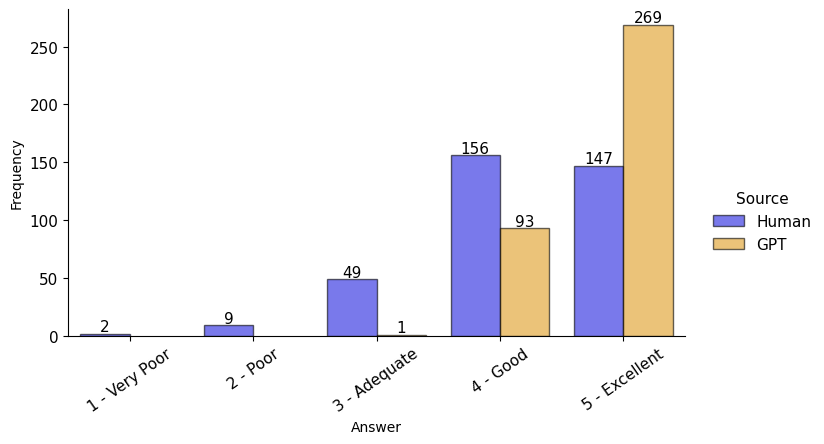}
	\caption{Fluency scores between human annotators and GPT-4o-mini}
	\label{fig:fluency}
\end{figure}

To capture the linguistic quality of the generated claims, we compare fluency scores between both annotators. \autoref{fig:fluency} shows the distribution of the fluency scores between human annotators and GPT-4o-mini. Both annotators rated the majority of claims as either "Good" or "Excellent", reflecting the quality of our generated claims. However, human annotators have shown to rate 49 of our generated claims as "Adequate" where GPT-4o-mini only answered 2 for this category. This indicates that some of these structures for these claims do not show human-like fluency, while they do show that to GPT-4o-mini. Meanwhile, GPT-4o-mini rated 231 claims as "Excellent" when compared to human annotators at 147. We suspect that the GPT-4o-mini is biased toward good answers, unlike human annotators who showed some restrictions when assessing these claims.

\begin{figure}[!ht]
	\centering
	\includegraphics[width=\linewidth]{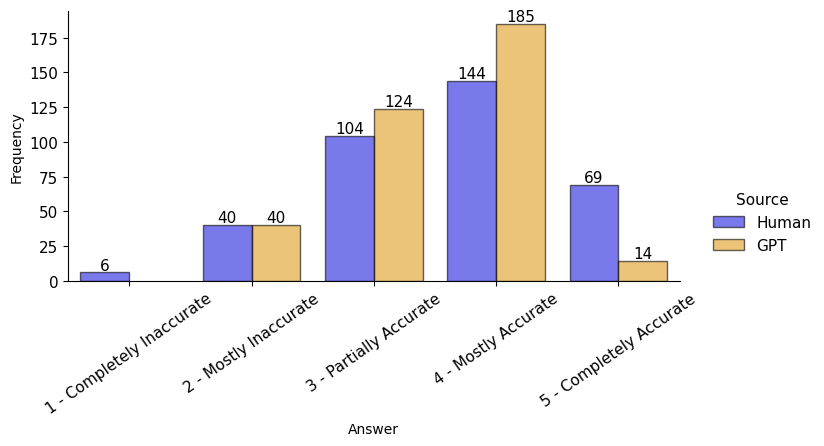}
	\caption{Factuality scores between human annotators and GPT-4o-mini}
	\label{fig:factuality}
\end{figure}

Furthermore, we assess factuality ratings across all annotators to understand the factuality of the generated claims. \autoref{fig:factuality} presents the distribution of factuality scores between GPT-4o-mini and human annotators. Both sources generally agree that our claims are factually accurate as indicated by the high "Mostly Accurate" and "Completely Accurate" counts. A notable trend here is that both annotators have rated 69 claims to be "Completely Accurate" while GPT-4o-mini rated 25 claims for the same category. This indicates that humans may have relied on their personal judgment to assess the factuality of these generated claims as opposed to GPT-4o-mini which uses its parametric knowledge. Despite this divergence, the overall similarity in distribution across the scale reflects a broad agreement between human and GPT-4o-mini.

\begin{figure}[!ht]
	\centering
	\includegraphics[width=\linewidth]{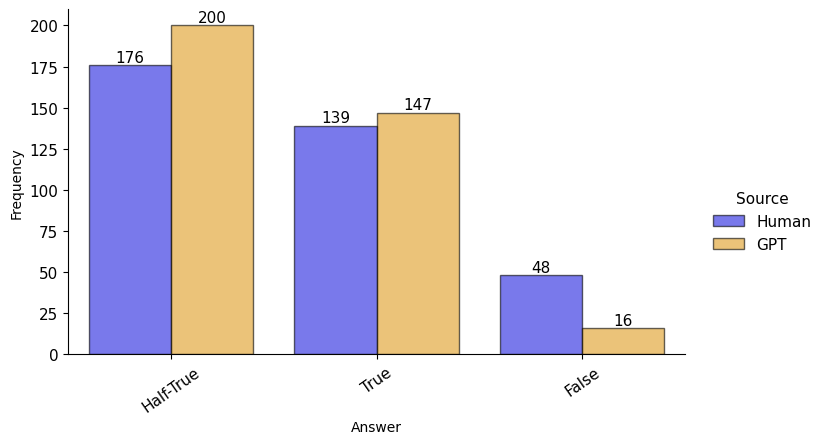}
	\caption{Label assignment between human annotators and GPT-4o-mini}
	\label{fig:label_assignment}
\end{figure}

Lastly, to examine alignment in veracity classification, we compare label assignments between human annotators and GPT-4o-mini. \autoref{fig:label_assignment} shows the label assignment between humans and GPT-4o-mini. Both annotators labeled a large distribution of claims as "Half-True" and "True", indicating similar agreement in their assessments. A notable trend is the high 48 "False" count from human annotators when compared to GPT-4o-mini at 25. This disparity may indicate that human annotators applied more precise or stringent criteria when identifying factual inaccuracies, unlike GPT-4o-mini which may have been more lenient.
\end{document}